\documentclass{article}

\usepackage[a4paper,margin=1in]{geometry}
\usepackage{times}
\usepackage[numbers,sort&compress]{natbib}

\usepackage{amsmath}
\usepackage{amsfonts}
\usepackage{amssymb}
\usepackage{bm}
\usepackage{graphicx}
\usepackage{booktabs}
\usepackage{multirow}
\usepackage{makecell}
\usepackage{pifont}
\usepackage{xcolor}
\usepackage{xspace}
\usepackage{hyperref}
\usepackage{url}

\newcommand{\sys}{WxFM-XL\xspace}

\newcommand{\eg}{\emph{e.g.}\xspace}
\newcommand{\ie}{\emph{i.e.}\xspace}

\usepackage{enumitem}

\newcommand{\myparagraph}[1]{\par\noindent\textbf{#1}\hspace{0.5em}\ignorespaces}
\usepackage{array}

\definecolor{bestcolor}{HTML}{E64B35}
\definecolor{secondcolor}{HTML}{3C5488}

\newcommand{\best}[1]{\textbf{#1}}
\newcommand{\second}[1]{\underline{#1}}

\title{WxFM-XL: Adapting Univariate Foundation Models to Multi-Station Weather Forecasting}

\author{Xiao Wang \quad Changjian Chen \quad Zhuo Tang \\
Rongwen Li \quad Hongwu Liu \quad Kenli Li}

\date{}

\begin{document}

\maketitle

\begin{abstract}
With the rise of univariate time series foundation models (\eg, Sundial, Timer), initial efforts have been made to extend them to multivariate settings.
However, these models mainly focus on modeling correlations among variables.
When they are applied to multi-station weather forecasting, two important factors are often overlooked: (1) the spatial information of stations, and (2) different error priors of different stations relative to the foundation model.
In this paper, we propose {\sys}, a model for adapting univariate time series foundation models to multi-station weather forecasting.
{\sys} introduces a cross-station error correlation prior graph to capture stationwise error priors with respect to the foundation model. 
Building on this, we further propose a dynamic fusion mechanism that adaptively integrates a spatial correlation graph with the error correlation prior graph.
Experiments on multiple datasets demonstrate that our model outperforms state of the art baselines.
\end{abstract}

\section{Introduction}
\label{sec:intro}
Building on the recent progress of time series forecasting, time series foundation models have demonstrated strong temporal priors (\eg, Sundial \citep{liu2025sundial}, Timer \citep{liu2024timer}).
These foundation models learn from large-scale cross-domain data and produce strong univariate forecasts with limited fine tuning, sometimes even in zero-shot settings.
However, their performance is limited when directly applied to multi-station weather forecasting. 
This is because these foundation models are primarily designed for univariate modeling and fall short in capturing the spatiotemporal relationships across stations.

\begin{figure}[t]
    \centerline{\includegraphics[width=0.95\columnwidth]{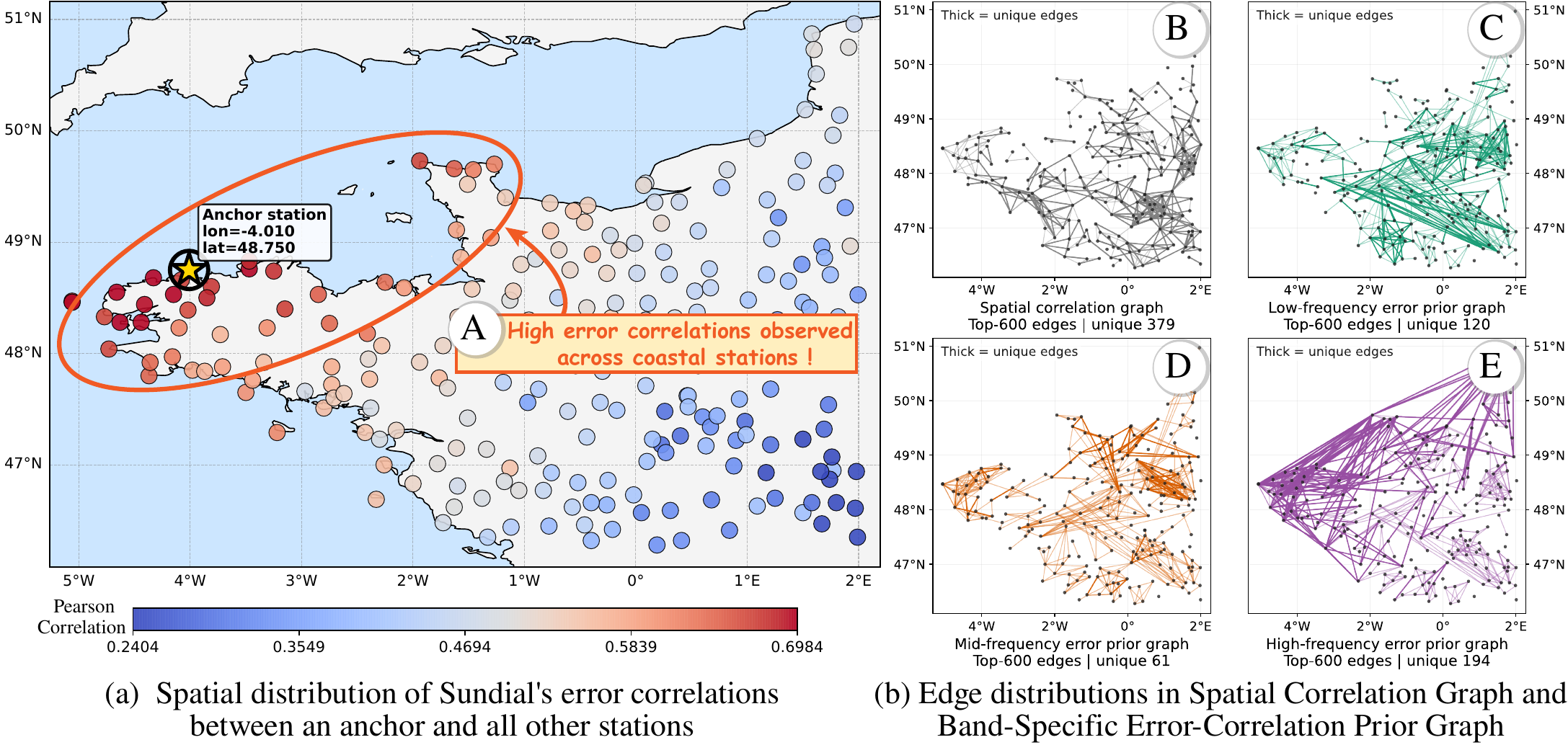}}
    \caption{Visualization of error correlation dependencies for the v-component of wind speed in the French region. (a) Sundial's forecast error correlations relative to an anchor station (star). (b) Topologies of spatial and band-specific error correlation graphs (showing top 600 edges with unique connections highlighted).}
    \label{fig:station_error_priors}
\end{figure}

To bridge this gap, some initial efforts have been made to extend univariate foundation models to multivariate settings, such as AdaPTS \citep{benechehab2025adapts} and Timer-XL \citep{liu2025timerxl}.
However, directly extending these multivariate methods to multi-station forecasting by treating each station as a variable leads to two limitations.
First, these models mainly extend foundation models to modeling correlations among multiple stations but ignore the spatial information of stations (\eg, neighborhood relationships).
Existing methods have shown that such spatial information is critical for multi-station forecasting \citep{wu2023Corrformer}.
Second, stations exhibit different error priors with respect to the foundation model. 
Recent studies have highlighted that a model's intrinsic errors serve as critical learning signals, implicitly reflecting how well the model captures underlying data structures \citep{liu2026beyond}.
For example, as shown in Fig.~\ref{fig:station_error_priors}(a), by calculating the error correlations between a coastal anchor station and all others, we observe complex error correlation priors that extend beyond simple neighborhood relationships.
Specifically, the anchor station shares strong error correlations with distant coastal stations (\eg, Fig.~\ref{fig:station_error_priors}A).
This indicates that relying solely on static station spatial relations is insufficient to capture these foundation model-specific error behaviors.

To address these limitations, rather than treating stations as variables, a method that can natively model the spatial information of stations and their different error priors is desired.
However, designing such a method is nontrivial. 
The spatial information of stations is inherently \textbf{static}, remaining constant regardless of the foundation model.
In contrast, the error priors are inherently \textbf{dynamic}.
These error patterns are not fixed.
Rather, they are inherently specific to the chosen foundation model and exhibit fluctuations across different frequency bands.
The static and dynamic natures of these two types of information make it challenging to jointly model and fuse them in a unified method, since the model must preserve time-invariant spatial dependencies while dynamically adapting to these complex error priors.
This inspires us to consider answering the following question in this study: \textbf{How can we effectively integrate static station spatial relations with dynamic error priors to improve forecasting performance?}

To address these challenges, we propose {\sys}, a model for adapting univariate time series foundation models to multi-station weather forecasting.
{\sys} first applies frequency domain decomposition to disentangle the foundation model forecasts into distinct frequency components.
Subsequently, tailored error correlation prior graphs are constructed for each specific frequency band.
As illustrated in Fig.~\ref{fig:station_error_priors}(b), these band-specific error priors exhibit distinct topological structures: while the static spatial correlation graph (Fig.~\ref{fig:station_error_priors}B) relies solely on geographic proximity, the error correlation prior graphs for low, mid, and high (Figs.~\ref{fig:station_error_priors}C-~\ref{fig:station_error_priors}E) frequencies reveal nonlocal connections that capture complex patterns beyond simple neighborhood relations (\eg, Fig.~\ref{fig:station_error_priors}A).
Building on this structural divergence, {\sys} introduces a dynamic fusion mechanism that adaptively integrates the spatial correlation graph with these band-specific error correlation prior graphs, resulting in frequency-aware dynamic fusion prior graphs.
Guided by these fusion graphs, we implement a lightweight graph residual adapter to extend the foundation model, which effectively injects both the spatial information of stations and the stationwise error priors under a weak fine tuning, while preserving the strong temporal priors of the foundation model.
Experiments across multiple datasets demonstrate that our model outperforms the SOTA (state of the art) baselines.

\noindent In summary, our contributions are as follows.
\begin{itemize}[leftmargin=*, itemsep=2pt, topsep=2pt, parsep=0pt, partopsep=0pt]
\item \textbf{A foundation model adaptation framework}, which bridges the gap between univariate time series foundation models and multi-station weather forecasting tasks.
\item \textbf{A dynamic fusion mechanism} that adaptively integrates static station spatial relations with dynamic error priors.
\item \textbf{Experiments on multiple datasets} demonstrate that our model outperforms SOTA baselines.
\end{itemize}

\section{Related Work}
\label{sec:related}
\myparagraph{Time series foundation models.}
Time series foundation models transfer temporal knowledge learned from large-scale, cross-domain corpora to downstream forecasting tasks.
For example, Chronos \citep{ansari2024chronos} quantizes continuous observations into discrete tokens and formulates forecasting as language modeling, whereas TimesFM \citep{das2024timesFM} adopts patch-based modeling to support zero-shot forecasting across diverse datasets.
Subsequent studies further improve scalability and generative capability.
Time-MoE \citep{shi2025timemoe} employs sparse mixture-of-experts to scale model capacity without proportionally increasing inference cost, while Timer \citep{liu2024timer} uses a GPT-style architecture and next-token prediction to learn from heterogeneous time series.
Sundial \citep{liu2025sundial} further introduces flow matching for probabilistic forecasting over continuous-valued observations, avoiding discrete tokenization.
Despite this progress, most foundation models are designed for univariate forecasting, limiting their performance in multi-station weather settings.

\myparagraph{Multi-station forecasting models.}
Multi-station forecasting methods differ mainly in how they represent station interactions and temporal evolution.
Graph-based methods represent stations as nodes and learn their spatial interactions.
StemGNN \citep{Cao2020StemGNN} jointly models inter-series correlations and temporal dynamics in the spectral domain, while MSGNet \citep{cai2024msgnet} extends this idea with multiscale frequency decomposition and adaptive graph convolution.
For long-range temporal modeling, Autoformer \citep{wu2021autoformer} combines seasonal trend decomposition with an Auto-Correlation mechanism, while FEDformer \citep{zhou2022fedformer} introduces frequency domain modeling to improve long horizon forecasting.
Corrformer \citep{wu2023Corrformer} further combines spatial cross-correlation with temporal autocorrelation for global and regional forecasting.
CDPNet \citep{xu2025CDPNet} instead models continuous spatiotemporal evolution from discrete station observations.
More recently, STELLA \citep{fu2025STELLA} replaces Transformer layers with a lightweight MLP to model spatiotemporal correlations without attention.
These methods demonstrate the importance of station dependencies, but they generally rely on task-specific architectures trained from scratch and therefore cannot directly exploit the temporal priors acquired through large-scale foundation model pretraining.

\myparagraph{Foundation models for multivariate forecasting.}
Recent studies bridge foundation models and multivariate forecasting through either native multivariate architectures or lightweight adaptation.
Moirai \citep{gerald2024moirai} develops a unified foundation model that supports heterogeneous frequencies and arbitrary numbers of variables.
Building on Timer, Timer-XL \citep{liu2025timerxl} extends next-token prediction to multivariate series and introduces causal TimeAttention to capture dependencies across variables and long contexts.
Rather than modifying the foundation model backbone, AdaPTS \citep{benechehab2025adapts} uses lightweight adapters to project multivariate inputs into latent representations that can be processed by existing univariate foundation models.
However, when applied to multi-station weather forecasting, these extensions often treat stations as generic variables and lack explicit modeling of the spatial information of stations or the correlation of stationwise error priors relative to the foundation model. 

\section{Problem Formulation}
\label{sec:problem}
In this section, we introduce the multi-station weather forecasting problem.
For clarity of presentation, we consider forecasting a single weather variable (\eg, temperature).
The formulation can be straightforwardly extended to multi-variable forecasting by applying the same procedure to each variable.
We study multi-station weather forecasting with $N$ weather stations distributed in a region.
Each station has a fixed geographic location (\eg, longitude, latitude, and elevation), and we denote the collection of station locations as $\mathbf{L}\in\mathbb{R}^{N\times 3}$.
At each timeframe $t$, a target weather variable at all stations is denoted by $\mathbf{x}_t \in \mathbb{R}^{N}$.
Given the historical observations over the past $T_{\mathrm{in}}$ timeframes $\mathbf{x}_{t-T_{\mathrm{in}}:t-1}$, the goal is to forecast the future weather conditions for the next $T_{\mathrm{out}}$ timeframes 
$\hat{\mathbf{x}}_{t:t+T_{\mathrm{out}}-1} = f\!\left(\mathbf{x}_{t-T_{\mathrm{in}}:t-1}, \mathbf{L}\right)$,
where $f(\cdot)$ denotes a multi-station forecasting model.

In this work, we focus on adapting a univariate time series foundation model to multi-station weather forecasting.
Let $f_\mathrm{FM}(\cdot)$ denote a time series foundation model that takes a univariate history and outputs a $T_{\mathrm{out}}$-step forecast for a single station.
Applying $f_\mathrm{FM}(\cdot)$ independently to $N$ weather stations yields stationwise foundation model forecasts $\tilde{\mathbf{x}}_{t: t+T_{\text {out }}-1}^{\mathrm{FM}} = f_\mathrm{FM}\!\left(\mathbf{x}_{t-T_{\mathrm{in}}:t-1}\right)$.
Our goal is to improve these independent stationwise forecasts by extending the foundation model with a dynamic fusion mechanism and a lightweight graph residual adapter.

\section{The {\sys} Framework}
\label{sec:methods}
\begin{figure}[t]
    \centering
    \includegraphics[width=\linewidth]{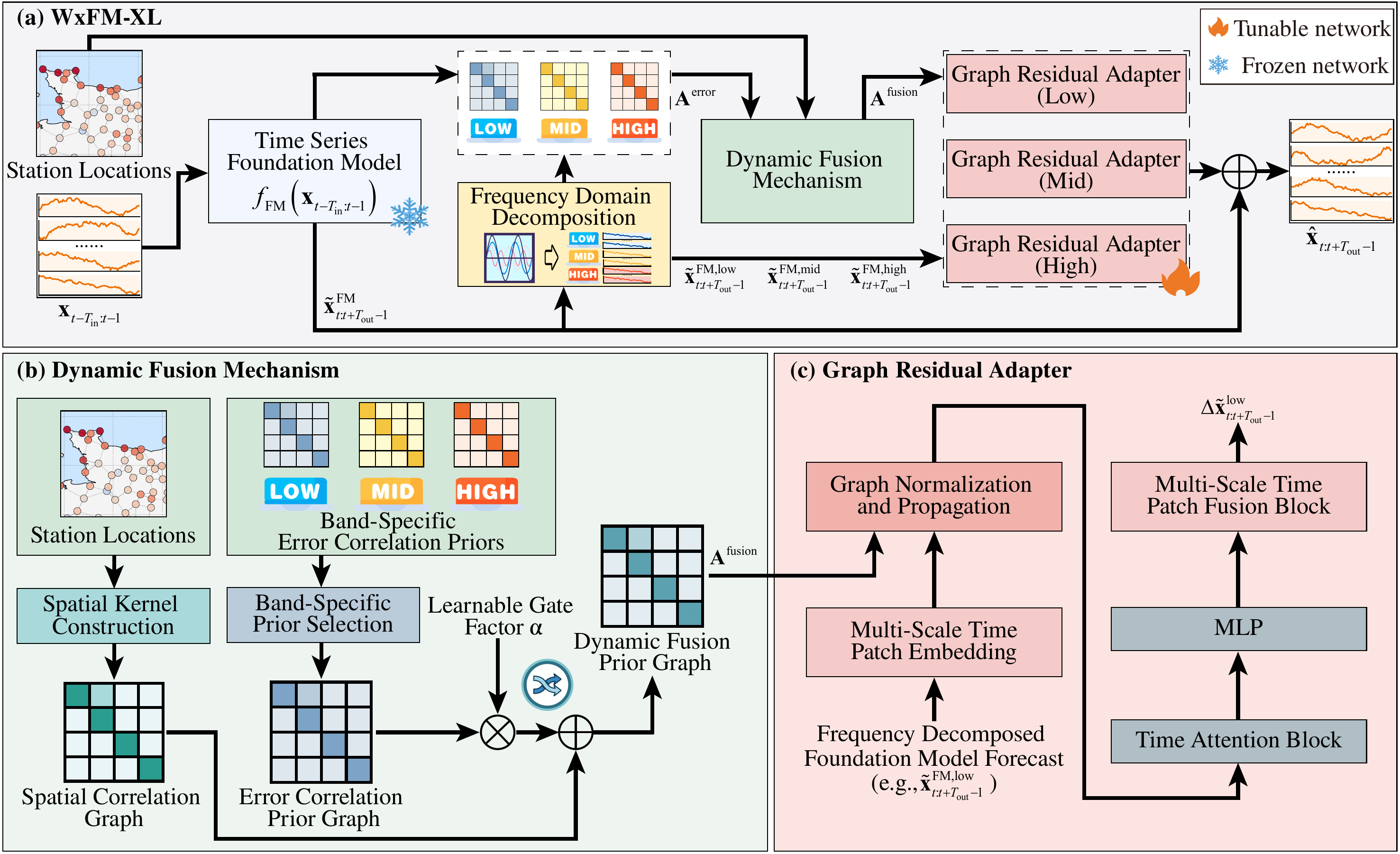}
    \caption{Overview of the {\sys} framework.}
    \label{fig:system}
\end{figure}
We introduce the {\sys} framework for adapting univariate time series foundation models to multi-station weather forecasting.
As shown in Fig.~\ref{fig:system}, {\sys} consists of two key components:
(1) a \textbf{Dynamic Fusion Mechanism} that constructs and combines the spatial correlation graph with error correlation prior graphs;
and (2) a \textbf{Graph Residual Adapter} that refines foundation model forecasts using the resulting dynamic fusion prior graph while preserving strong temporal priors.

\subsection{Dynamic Fusion Mechanism}
\label{sec:dfm}
As shown in Fig.~\ref{fig:system}(a), the Dynamic Fusion Mechanism first applies \textbf{(1) frequency domain decomposition} to decompose the foundation model forecasts into distinct frequency bands (\ie, low, mid, and high frequencies).
Then, we construct different \textbf{(2) error correlation prior graphs} for each specific frequency band, and the \textbf{(3) spatial correlation graph} based on the spatial information.
Finally, we develop a gated fusion unit to obtain the \textbf{(4) dynamic fusion prior graph}.

\myparagraph{(1) Frequency domain decomposition.}
Foundation model errors exhibit different patterns across frequency bands.
To isolate these patterns, we follow frequency domain modeling \citep{Yang2024FBM} and decompose the foundation model forecasts into low, mid, and high frequency components.
Let
$\tilde{\mathbf{x}}^{\mathrm{FM}}_{t:t+T_{\mathrm{out}}-1}
\in\mathbb{R}^{T_{\mathrm{out}}\times N}$
denote the foundation model forecasts and
$\mathcal{F}=\{\mathrm{low},\mathrm{mid},\mathrm{high}\}$
the set of frequency bands.
For each $f\in\mathcal{F}$, we retain the Fourier coefficients associated with the corresponding frequency band and reconstruct the component as
\begin{equation}
\label{eq:frequency_decomposition}
\tilde{\mathbf{x}}^{\mathrm{FM},f}_{t:t+T_{\mathrm{out}}-1}
=
\operatorname{IDFT}\!\left(
\mathcal{P}_f\!\left(
\operatorname{DFT}\!\left(
\tilde{\mathbf{x}}^{\mathrm{FM}}_{t:t+T_{\mathrm{out}}-1}
\right)
\right)
\right),
\qquad f\in\mathcal{F},
\end{equation}
where $\operatorname{DFT}(\cdot)$ transforms the forecasts into the frequency domain, $\mathcal{P}_f$ retains the Fourier coefficients assigned to frequency band $f$ and their conjugate counterparts, and $\operatorname{IDFT}(\cdot)$ reconstructs the corresponding component in the time domain.
The detailed decomposition and reconstruction are provided in Appendix~\ref{app:frequency_decomposition}.
These components are subsequently used to construct frequency dependent error correlation prior graphs.

\myparagraph{(2) Error correlation prior graph.}
Stationwise error correlations vary across frequency bands, such that the same station pair may exhibit different dependencies in different frequency components (\eg, Fig.~\ref{fig:station_error_priors}(b)).
Therefore, for each $f\in\mathcal{F}$, we construct an error correlation prior graph
$\mathbf{A}^{\mathrm{error}}_f\in\mathbb{R}^{N\times N}$.
We collect the forecast errors associated with frequency band $f$ over the training set and flatten the sample and lead time dimensions into
$\mathbf{E}^f\in\mathbb{R}^{M\times N}$,
where $M$ is the number of flattened instances and $E^f_{m,i}$ denotes the error at station $i$ for instance $m$.
The adjacency weight between stations $i$ and $j$ is defined by their Pearson correlation:
\begin{equation}
\label{eq:corr_graph}
\left(\mathbf{A}^{\mathrm{error}}_f\right)_{i,j}
=
\frac{
\sum_{m=1}^{M}
\left(E^f_{m,i}-\bar{E}^f_i\right)
\left(E^f_{m,j}-\bar{E}^f_j\right)
}{
\sqrt{
\sum_{m=1}^{M}
\left(E^f_{m,i}-\bar{E}^f_i\right)^2
}
\sqrt{
\sum_{m=1}^{M}
\left(E^f_{m,j}-\bar{E}^f_j\right)^2
}
},
\end{equation}
where
$\bar{E}^f_i=\frac{1}{M}\sum_{m=1}^{M}E^f_{m,i}$
is the mean error at station $i$.
We retain up to $k$ stations with the largest positive Pearson correlations, set non-positive and unselected weights to zero, symmetrize the graph, and set self-loop weights to one.

\myparagraph{(3) Spatial correlation graph.}
We construct a spatial correlation graph from the fixed geographic locations of stations.
Let
$\mathbf{L}=[\mathbf{l}_1,\ldots,\mathbf{l}_N]^\top
\in\mathbb{R}^{N\times 3}$
denote the longitude, latitude, and elevation of the stations.
After standardizing the station locations, we construct the spatial adjacency matrix using a Gaussian kernel:
\begin{equation}
\label{eq:spatial_graph}
\left(\mathbf{A}^{\mathrm{spatial}}\right)_{i,j}
=
\exp\!\left(
-\frac{
\left\lVert
\tilde{\mathbf{l}}_i-\tilde{\mathbf{l}}_j
\right\rVert_2^2
}{
2\sigma^2
}
\right),
\end{equation}
where $\tilde{\mathbf{l}}_i$ is the standardized location of station $i$, and $\sigma$ is the median of the nonzero pairwise distances.
We retain the $k$ nearest neighbors of each station, including self-loops, set the remaining entries to zero, and symmetrize $\mathbf{A}^{\mathrm{spatial}}$.
Since $\mathbf{A}^{\mathrm{spatial}}$ depends only on $\mathbf{L}$, it is static across time.

\myparagraph{(4) Dynamic fusion prior graph.}
As shown in Fig.~\ref{fig:system}(b), for each frequency band $f\in\mathcal{F}$, we combine the error correlation prior graph $\mathbf{A}^{\mathrm{error}}_f$ with the spatial correlation graph $\mathbf{A}^{\mathrm{spatial}}$ using a learnable gate $\alpha_f\in\mathbb{R}$:
\begin{equation}
\label{eq:dynamic_fusion_prior}
\mathbf{A}^{\mathrm{fusion}}_f
=
\mathbf{A}^{\mathrm{spatial}}
+
\alpha_f\mathbf{A}^{\mathrm{error}}_f.
\end{equation}
The spatial graph encodes geographic relations that remain fixed across foundation models, whereas the error graph captures dependencies specific to the selected foundation model and frequency band.
By learning $\alpha_f$ for each band, the fusion mechanism balances these static and dynamic priors, preserving geographic constraints while incorporating nonlocal error relations beyond spatial proximity.
The resulting graph guides the Graph Residual Adapter in refining
$\tilde{\mathbf{x}}^{\mathrm{FM},f}_{t:t+T_{\mathrm{out}}-1}$
according to the error structure of each frequency component.

\subsection{Graph Residual Adapter}
\label{sec:adapter}
For each frequency band $f\in\mathcal{F}$, we employ an independent lightweight Graph Residual Adapter to refine
$\tilde{\mathbf{x}}^{\mathrm{FM},f}_{t:t+T_{\mathrm{out}}-1}$
under the corresponding dynamic fusion prior graph
$\mathbf{A}^{\mathrm{fusion}}_f$,
while keeping the foundation model frozen.
As shown in Fig.~\ref{fig:system}(c), we first process the input with the \textbf{(1) multi-scale time patch embedding}.
Then, we perform the \textbf{(2) graph normalization and propagation} to aggregate information on the dynamic fusion prior graph.
Finally, the \textbf{(3) multi-scale patch fusion block} combines representations from different temporal scales.
All these modules are trained in an \textbf{(4) end-to-end training} manner.

\myparagraph{(1) Multiscale time patch embedding.}
Since the frozen foundation model already provides strong temporal priors, we use time patching to efficiently extract local residual patterns.
For each patch size $p\in\mathcal{P}$ (\eg, $\mathcal{P}=\{6,12,24\}$), the frequency component
$\tilde{\mathbf{x}}^{\mathrm{FM},f}_{t:t+T_{\mathrm{out}}-1}
\in\mathbb{R}^{T_{\mathrm{out}}\times N}$
is divided into nonoverlapping temporal segments and projected into latent tokens:
\begin{equation}
\label{eq:patch_embedding}
\mathbf{Z}^{(p)}
=
\operatorname{Reshape}_{p}\!\left(
\tilde{\mathbf{x}}^{\mathrm{FM},f}_{t:t+T_{\mathrm{out}}-1}
\right)
\mathbf{W}_{p}
+
\mathbf{b}_{p}.
\end{equation}
Here,
$\operatorname{Reshape}_{p}(\cdot)$ produces
$T_p=T_{\mathrm{out}}/p$ patches with shape
$\mathbb{R}^{N\times T_p\times p}$,
while
$\mathbf{W}_{p}\in\mathbb{R}^{p\times D}$
and
$\mathbf{b}_{p}\in\mathbb{R}^{D}$
are learnable parameters.
The resulting representation
$\mathbf{Z}^{(p)}\in\mathbb{R}^{N\times T_p\times D}$
is passed to the graph propagation module.
Using multiple patch sizes enables the adapter to capture residual patterns at different temporal granularities.

\myparagraph{(2) Graph normalization and propagation.}
Given the frequency-specific dynamic fusion prior graph
$\mathbf{A}^{\mathrm{fusion}}_f$, we use a learnable node gate
$\mathbf{g}\in\mathbb{R}^{N}$ to adapt its propagation weights.
For each patch size $p\in\mathcal{P}$, graph normalization and propagation are defined as
\begin{equation}
\label{eq:adapter_graphprop}
\begin{aligned}
\widetilde{\mathbf{A}}^{\mathrm{fusion}}_f
&=
\mathrm{RowNorm}\!\left(
\mathbf{A}^{\mathrm{fusion}}_f
\odot
\mathrm{sig}\!\left(
\mathbf{g}\mathbf{1}^{\top}
+
\mathbf{1}\mathbf{g}^{\top}
\right)
\right),\\
\widetilde{\mathbf{Z}}^{(p)}
&=
\widetilde{\mathbf{A}}^{\mathrm{fusion}}_f
\operatorname{TokenNorm}\!\left(\mathbf{Z}^{(p)}\right),
\qquad p\in\mathcal{P}.
\end{aligned}
\end{equation}
Here, $\mathrm{sig}(\cdot)$ denotes the sigmoid function, and
$\mathrm{RowNorm}(\cdot)$ normalizes each row to sum to one.
The multiplication is performed along the station dimension, producing
$\widetilde{\mathbf{Z}}^{(p)}
\in\mathbb{R}^{N\times T_p\times D}$ by aggregating residual information
across stations. These graph-propagated representations are then used for
temporal refinement and residual reconstruction.

\myparagraph{(3) Multi-scale patch fusion block.}
After graph propagation, we obtain the token representation
$\widetilde{\mathbf{Z}}^{(p)}$ for each patch size
$p\in\mathcal{P}$.
We further refine these representations using a lightweight temporal module
consisting of a Time Attention block followed by the MLP. This module captures interactions along the temporal token dimension
while keeping the adapter lightweight.
The refined representations are mapped back to the temporal domain through
the corresponding patch unembedding operation, producing a residual forecast
$\Delta\tilde{\mathbf{x}}^{(p,f)}_{t:t+T_{\mathrm{out}}-1}
\in\mathbb{R}^{T_{\mathrm{out}}\times N}$
for patch size $p$ and frequency band $f$.
Details of the temporal module, patch unembedding, and scale weight scorer
are provided in Appendix~\ref{app:adapter_details}.

To combine residuals obtained from different patch sizes, we compute
normalized, nonnegative, and sample-dependent weights $\theta^{(p)}$ and
perform weighted fusion:
\begin{equation}
\label{eq:ms_fusion}
\Delta\tilde{\mathbf{x}}^{f}_{t:t+T_{\mathrm{out}}-1}
=
\sum_{p\in\mathcal{P}}
\theta^{(p)}
\Delta\tilde{\mathbf{x}}^{(p,f)}_{t:t+T_{\mathrm{out}}-1}.
\end{equation}

Finally, the adapter combines the residual corrections from all frequency
bands. A learnable factor $\beta_f\in\mathbb{R}$ controls the contribution
of each band:
\begin{equation}
\label{eq:final_forecast}
\hat{\mathbf{x}}_{t:t+T_{\mathrm{out}}-1}
=
\sum_{f\in\mathcal{F}}
\left(
\tilde{\mathbf{x}}^{\mathrm{FM},f}_{t:t+T_{\mathrm{out}}-1}
+
\beta_f
\Delta\tilde{\mathbf{x}}^{f}_{t:t+T_{\mathrm{out}}-1}
\right).
\end{equation}

\myparagraph{(4) End-to-end training.}
We keep the foundation model frozen and jointly optimize all adapter parameters,
including the node gates $\mathbf{g}$, prior fusion factors $\alpha_f$, and
band scaling factors $\beta_f$.
Because low-frequency components usually contain substantially more energy,
a standard forecasting loss may underemphasize errors in high-frequency
components. We therefore combine the MSE of the final forecast with an
energy-normalized RMSE for each frequency band.

For clarity, we omit the forecast-window subscripts and define
$\hat{\mathbf{x}}^{f}
=
\tilde{\mathbf{x}}^{\mathrm{FM},f}
+
\beta_f\Delta\tilde{\mathbf{x}}^{f}$.
The training objective is
\begin{equation}
\label{eq:training_loss}
\mathcal{L}
=
\lambda_{\mathrm{mse}}
\mathrm{MSE}\!\left(\hat{\mathbf{x}},\mathbf{x}\right)
+
\sum_{f\in\mathcal{F}}
\lambda_f
\sqrt{
\frac{
\left\lVert
\hat{\mathbf{x}}^{f}-\mathbf{x}^{f}
\right\rVert_{\mathrm{F}}^2
}{
\left\lVert
\mathbf{x}^{f}
\right\rVert_{\mathrm{F}}^2+\varepsilon
}
}.
\end{equation}
Here, $\mathbf{x}^{f}$ denotes the ground-truth component obtained using the
same frequency decomposition, $\varepsilon>0$ ensures numerical stability,
and $\lambda_{\mathrm{mse}}$ and $\lambda_f$ balance the two loss terms.
Energy normalization prevents low-energy frequency components from being
overwhelmed during training, enabling balanced residual correction across
frequency bands.

\section{Experiments}
\label{sec:eval}
\newcommand{\std}[1]{\scalebox{0.68}{$_{\pm #1}$}}
\newcommand{\nostdval}[1]{#1}

\begin{table}[t]
\centering
\caption{MSE results on the Hunan regional, French regional, and global
datasets with a 96-hour input window and forecast horizons of 24 and
96 hours. Best and second-best results are shown in
\best{bold} and \second{underlined}, respectively, based on unrounded
values. Complete MAE results are reported in the appendix.}
\label{tab:all_mse}

\begingroup
\scriptsize
\renewcommand{\arraystretch}{1.9}
\setlength{\tabcolsep}{0.5pt}
\resizebox{\linewidth}{!}{%
\begin{tabular}{@{}c|c|c|cc!{\vrule width 0.8pt}cccccccccccc@{}}
\hline
\multicolumn{3}{c|}{Models}
& \makecell{\sys\\(Timer)}
& \makecell{\sys\\(Sundial)}
& Timer
& Sundial
& \makecell{AdaPTS\\(Timer)}
& \makecell{AdaPTS\\(Sundial)}
& Moirai$_{\mathrm{ft}}$
& Moirai
& Timer-XL$_{\mathrm{ft}}$
& Timer-XL
& STELLA
& CDPNet
& MSGNet
& Corrformer \\
\hline

\multirow{6}{*}{\rotatebox{90}{Hunan}}
& \multirow{2}{*}{\rotatebox{90}{temp}}
& 24
& \second{5.14}\std{0.01}
& \best{5.06}\std{0.09}
& \nostdval{5.28}
& \nostdval{5.78}
& 5.33\std{0.05}
& 5.57\std{0.11}
& 5.65\std{0.27}
& \nostdval{5.24}
& 5.43\std{0.15}
& \nostdval{5.26}
& 8.56\std{0.84}
& 7.44\std{0.79}
& 5.89\std{0.39}
& 6.01\std{0.47} \\
&
& 96
& \best{7.31}\std{0.04}
& \second{7.33}\std{0.05}
& \nostdval{7.52}
& \nostdval{8.35}
& 7.73\std{0.42}
& 8.51\std{0.51}
& 7.54\std{0.29}
& \nostdval{7.52}
& 9.29\std{0.27}
& \nostdval{7.37}
& 16.06\std{1.97}
& 26.94\std{6.19}
& 12.45\std{0.93}
& 8.60\std{0.48} \\
\cline{2-17}

& \multirow{2}{*}{\rotatebox{90}{v-wind}}
& 24
& \second{2.45}\std{0.03}
& \best{2.43}\std{0.01}
& \nostdval{2.63}
& \nostdval{3.07}
& 2.52\std{0.03}
& 2.54\std{0.14}
& 2.76\std{0.02}
& \nostdval{2.81}
& 2.58\std{0.01}
& \nostdval{2.58}
& 2.50\std{0.03}
& 2.63\std{0.09}
& 2.69\std{0.07}
& 3.46\std{0.12} \\
&
& 96
& \best{3.10}\std{0.00}
& 3.30\std{0.08}
& \nostdval{\second{3.12}}
& \nostdval{3.66}
& 3.23\std{0.11}
& 3.27\std{0.16}
& 3.62\std{0.32}
& \nostdval{3.83}
& 3.37\std{0.37}
& \nostdval{3.20}
& 3.43\std{0.34}
& 4.54\std{1.41}
& 4.62\std{0.28}
& 3.95\std{0.09} \\
\cline{2-17}

& \multirow{2}{*}{\rotatebox{90}{u-wind}}
& 24
& 2.08\std{0.00}
& \best{2.02}\std{0.00}
& \nostdval{2.16}
& \nostdval{2.57}
& 2.14\std{0.13}
& 2.16\std{0.02}
& 2.38\std{0.01}
& \nostdval{2.37}
& 2.14\std{0.01}
& \nostdval{2.16}
& \second{2.03}\std{0.01}
& 2.15\std{0.03}
& 2.30\std{0.12}
& 2.14\std{0.05} \\
&
& 96
& \second{2.15}\std{0.03}
& \best{2.15}\std{0.11}
& \nostdval{2.35}
& \nostdval{2.80}
& 2.31\std{0.04}
& 2.29\std{0.13}
& 2.71\std{0.03}
& \nostdval{2.82}
& 2.27\std{0.03}
& \nostdval{2.35}
& 2.21\std{0.00}
& 2.51\std{0.09}
& 2.89\std{0.15}
& 2.39\std{0.19} \\
\hline

\multirow{6}{*}{\rotatebox{90}{French}}
& \multirow{2}{*}{\rotatebox{90}{temp}}
& 24
& \second{6.57}\std{0.07}
& \best{6.50}\std{0.10}
& \nostdval{7.57}
& \nostdval{8.02}
& 7.29\std{0.21}
& 7.31\std{0.15}
& 7.23\std{0.05}
& \nostdval{7.51}
& 6.69\std{0.32}
& \nostdval{7.07}
& 8.36\std{0.21}
& 22.42\std{3.87}
& 6.98\std{0.35}
& 9.81\std{0.57} \\
&
& 96
& \best{12.08}\std{0.07}
& 12.54\std{0.39}
& \nostdval{\second{12.35}}
& \nostdval{13.52}
& 12.92\std{0.74}
& 13.52\std{0.43}
& 12.73\std{1.23}
& \nostdval{12.85}
& 14.17\std{2.14}
& \nostdval{12.52}
& 19.22\std{2.20}
& 83.78\std{3.77}
& 13.14\std{0.41}
& 13.37\std{0.51} \\
\cline{2-17}

& \multirow{2}{*}{\rotatebox{90}{v-wind}}
& 24
& \best{4.50}\std{0.03}
& \second{4.54}\std{0.01}
& \nostdval{5.69}
& \nostdval{6.11}
& 4.87\std{0.21}
& 5.24\std{0.23}
& 5.39\std{0.14}
& \nostdval{6.30}
& 6.01\std{0.05}
& \nostdval{5.66}
& 5.28\std{0.03}
& 5.15\std{0.35}
& 5.31\std{0.14}
& 6.06\std{0.39} \\
&
& 96
& \best{6.39}\std{0.04}
& \second{6.60}\std{0.28}
& \nostdval{7.96}
& \nostdval{9.08}
& 6.98\std{0.42}
& 7.18\std{0.37}
& 8.15\std{0.37}
& \nostdval{10.80}
& 8.46\std{0.29}
& \nostdval{7.91}
& 7.21\std{0.24}
& 7.83\std{0.88}
& 8.95\std{0.59}
& 8.14\std{0.50} \\
\cline{2-17}

& \multirow{2}{*}{\rotatebox{90}{u-wind}}
& 24
& \best{4.60}\std{0.03}
& \second{4.63}\std{0.01}
& \nostdval{6.26}
& \nostdval{7.10}
& 5.33\std{0.21}
& 5.70\std{0.30}
& 5.65\std{0.07}
& \nostdval{6.17}
& 4.90\std{0.06}
& \nostdval{6.54}
& 5.79\std{0.16}
& 5.36\std{1.07}
& 5.09\std{0.15}
& 6.08\std{0.40} \\
&
& 96
& \second{8.21}\std{0.13}
& \best{8.11}\std{0.14}
& \nostdval{9.79}
& \nostdval{11.45}
& 8.29\std{0.31}
& 8.60\std{0.39}
& 10.11\std{0.53}
& \nostdval{12.40}
& 8.81\std{0.34}
& \nostdval{9.94}
& 8.85\std{0.44}
& 9.28\std{0.19}
& 9.71\std{0.72}
& 9.51\std{0.26} \\
\hline

\multirow{4}{*}{\rotatebox{90}{Global}}
& \multirow{2}{*}{\rotatebox{90}{temp}}
& 24
& \second{7.57}\std{0.01}
& \best{7.49}\std{0.03}
& \nostdval{9.67}
& \nostdval{10.27}
& 8.73\std{0.52}
& 9.15\std{0.17}
& 9.07\std{0.06}
& \nostdval{9.87}
& 8.04\std{0.60}
& \nostdval{10.90}
& 7.61\std{0.01}
& 8.81\std{0.19}
& 9.90\std{0.31}
& 7.79\std{0.46} \\
&
& 96
& \best{13.95}\std{0.08}
& \second{13.99}\std{0.01}
& \nostdval{16.47}
& \nostdval{17.94}
& 15.77\std{0.45}
& 16.15\std{0.17}
& 16.24\std{0.07}
& \nostdval{18.55}
& 14.36\std{0.14}
& \nostdval{18.11}
& 14.46\std{0.18}
& 14.93\std{2.44}
& 17.97\std{1.35}
& 16.34\std{0.36} \\
\cline{2-17}

& \multirow{2}{*}{\rotatebox{90}{wind}}
& 24
& \best{3.58}\std{0.01}
& \second{3.66}\std{0.00}
& \nostdval{4.11}
& \nostdval{4.91}
& 4.03\std{0.15}
& 4.17\std{0.12}
& 4.24\std{0.02}
& \nostdval{4.83}
& 3.76\std{0.06}
& \nostdval{4.18}
& 3.73\std{0.01}
& 4.17\std{0.07}
& 5.25\std{0.39}
& 3.74\std{0.17} \\
&
& 96
& \best{4.69}\std{0.10}
& \second{4.76}\std{0.02}
& \nostdval{5.26}
& \nostdval{6.23}
& 4.81\std{0.22}
& 4.94\std{0.15}
& 5.42\std{0.03}
& \nostdval{7.36}
& 4.79\std{0.01}
& \nostdval{5.23}
& 4.85\std{0.02}
& 4.98\std{0.14}
& 6.97\std{0.06}
& 5.14\std{0.18} \\
\hline
\end{tabular}%
}
\endgroup
\vspace{-4.8mm}
\end{table}
\subsection{Experimental Settings}
\myparagraph{Datasets.}
We evaluate all methods on three hourly weather station datasets.
The MeteoNet dataset \citep{meteonet2020}, provided by the French national weather service, contains
observations from $234$ stations between January 1 and December 31, 2016,
including the u-component and v-component of wind at 10 meters and temperature
at 2 meters.
The Hunan dataset, provided by the Hunan Province Meteorological Bureau,
contains the same three variables from $3{,}570$ stations between April 1 and
September 30, 2023.
The global dataset, obtained from the National Centers for Environmental
Information (NCEI) \citep{noaa2026,wu2023Corrformer}, contains hourly averaged wind speed and
temperature from $3{,}850$ stations between January 1, 2019, and December 31,
2020.
For all datasets, we chronologically split the observations into training,
validation, and test sets using a $6{:}1{:}3$ ratio.
Detailed dataset information is provided in
Appendix~\ref{app:dataset_details}.

\myparagraph{Baselines.}
We compare {\sys} with three groups of forecasting methods.
\textbf{(1) Time series foundation models.}
Timer \citep{liu2024timer} and Sundial \citep{liu2025sundial} are evaluated
under zero-shot settings without parameter updates on the target datasets.
\textbf{(2) Multi-station forecasting models.}
STELLA \citep{fu2025STELLA}, CDPNet \citep{xu2025CDPNet}, MSGNet \citep{cai2024msgnet}, and
Corrformer \citep{wu2023Corrformer} are included as models that explicitly
capture dependencies across stations.
\textbf{(3) Foundation models for multivariate forecasting.}
Timer-XL \citep{liu2025timerxl} and Moirai \citep{gerald2024moirai} are
evaluated under both zero-shot and fine-tuning settings, with their
fine-tuned variants denoted as Timer-XL$_{\mathrm{ft}}$ and Moirai$_{\mathrm{ft}}$.
AdaPTS \citep{benechehab2025adapts} is combined with Timer and Sundial,
yielding AdaPTS (Timer) and AdaPTS (Sundial), respectively.
Detailed baseline information is provided in
Appendix~\ref{app:baseline_details}.

\myparagraph{Implementation details.}
All baselines follow their official implementations and recommended
settings. We use frozen Timer and Sundial backbones, denoted as
{\sys} (Timer) and {\sys} (Sundial). To accommodate Timer
\citep{liu2024timer}, all models use a $96$-hour input window, with
forecast horizons $T_{\mathrm{out}}\in\{24,96\}$.
We train {\sys} with Adam for up to 20 epochs, using a cosine learning
rate schedule from $10^{-4}$ to $10^{-6}$.

Stochastic experiments use three seeds (2021, 2022, and 2023); for
trainable methods, we report the mean and sample standard deviation.
Timer and Timer-XL use deterministic inference, while Sundial and
Moirai show negligible variation at the reported precision, so their
standard deviations are omitted. {\sys} uses three frequency bands
$\mathcal{F}=\{\mathrm{low},\mathrm{mid},\mathrm{high}\}$ and patch sizes
$\mathcal{P}=\{6,12,24\}$. Further settings are provided in
Appendix~\ref{app:implementation_details}.

\begin{table}[t]
\centering
\caption{Ablation results (MSE) on the Hunan regional and French regional
datasets with a 96-hour input window and a 24-hour forecast horizon.
Results are reported as the mean and sample standard deviation over three
random seeds. Best results are shown in \textbf{bold}. Complete MAE
results are provided in the appendix.}
\label{tab:ablation}

\begingroup
\small
\renewcommand{\arraystretch}{1}
\setlength{\tabcolsep}{3pt}

\begin{tabular}{@{}lcccccc@{}}
\toprule
\multirow{2}{*}{Model variant}
& \multicolumn{3}{c}{Hunan regional}
& \multicolumn{3}{c}{French regional} \\
\cmidrule(lr){2-4}
\cmidrule(lr){5-7}
& temp & v-wind & u-wind
& temp & v-wind & u-wind \\
\midrule

\textbf{\sys{} (Timer)}
& \best{5.143}\std{0.013}
& \best{2.450}\std{0.034}
& \best{2.084}\std{0.003}
& \best{6.571}\std{0.065}
& \best{4.501}\std{0.032}
& \best{4.602}\std{0.028} \\

\midrule

w/o frequency domain decomposition
& 5.393\std{0.022}
& 2.559\std{0.067}
& 2.140\std{0.011}
& 6.773\std{0.065}
& 4.696\std{0.032}
& 4.900\std{0.024} \\

w/o error correlation prior graph
& 5.430\std{0.033}
& 2.534\std{0.019}
& 2.110\std{0.002}
& 6.808\std{0.124}
& 4.672\std{0.073}
& 4.882\std{0.025} \\

w/o multi-scale temporal module
& 5.671\std{0.036}
& 2.616\std{0.035}
& 2.149\std{0.021}
& 6.986\std{0.108}
& 5.251\std{0.056}
& 5.551\std{0.035} \\

\bottomrule
\end{tabular}

\endgroup
\end{table}

\subsection{Result Analysis}
From the results in Table~\ref{tab:all_mse}, we draw three observations.

(1) \textbf{Overall performance.}
Across all 16 dataset-variable-horizon settings,
a {\sys} variant achieves the best MSE:
{\sys} (Timer) ranks first in nine settings and
{\sys} (Sundial) in the remaining seven.
Both variants consistently outperform their frozen backbones,
demonstrating the effectiveness of {\sys} across different regions,
variables, and forecasting horizons.

(2) \textbf{Comparison with task-specific models.}
Compared with STELLA, CDPNet, MSGNet, and Corrformer, {\sys} generally
achieves better performance.
On the four French wind forecasting settings, the best {\sys} variant
reduces MSE by $8.3\%$--$12.6\%$ relative to the strongest task-specific
baseline.
These results indicate that combining pretrained temporal priors with
spatial relations and frequency-specific error correlations provides
more effective cross-station modeling than task-specific architectures
trained from scratch.

(3) \textbf{Comparison with foundation model extensions.}
Fine tuning improves the average performance of both Timer-XL and
Moirai, but the gains are not uniform across forecasting tasks.
For example, Timer-XL fine tuning increases the
French 24-hour v-wind MSE from $5.66$ to $6.01$ and the Hunan 96-hour
temperature MSE from $7.37$ to $9.29$.
This variation indicates that generic multivariate fine tuning does not
uniformly adapt to the heterogeneous dependencies among weather stations.
AdaPTS also generally underperforms
{\sys}, indicating that generic multivariate adaptation is less reliable
than the proposed station-aware, frequency-specific residual correction.

\begin{figure}[t]
    \centering
    \includegraphics[width=\linewidth]{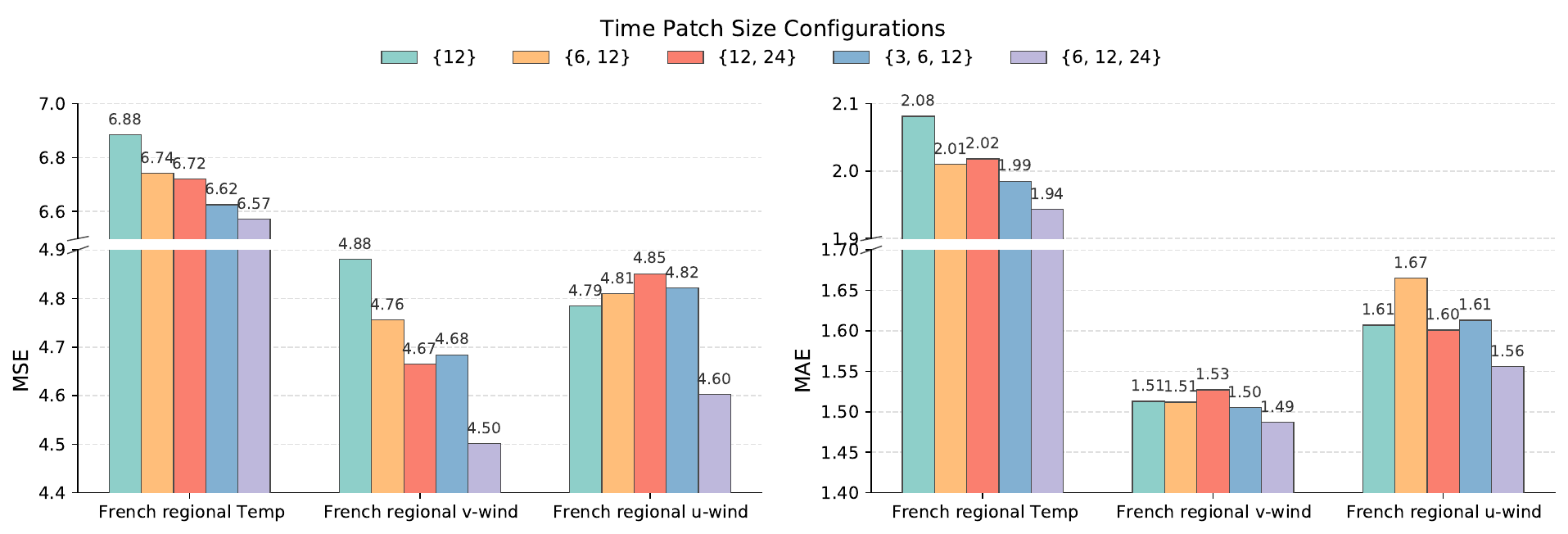}
    \caption{Performance comparison of {\sys} (Timer) under different time patch size configurations. The look-back window is set to 96, and the prediction horizon is 24.}
    \label{fig:patch_ablation}
    \vspace{-2.8mm}
\end{figure}

\begin{figure}[t]
    \centerline{\includegraphics[width=\linewidth]{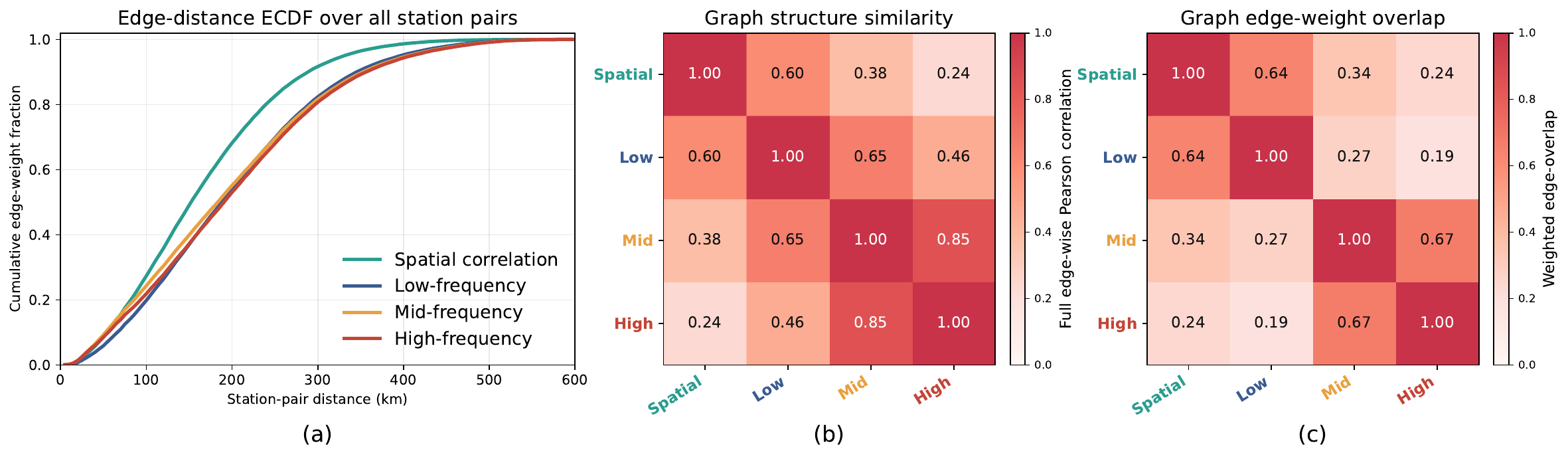}}
    \caption{Structural properties of prior graphs (taking the v-component of wind speed in the French region as an example).}
    \label{fig:full_edge_analysis}
    \vspace{-2.2mm}
\end{figure}

\subsection{Ablation Study}
We conduct component ablations on the Hunan and French regional datasets to evaluate the core designs of {\sys}.
Table~\ref{tab:ablation} compares the full model using Timer with three ablated variants.

\myparagraph{Impact of frequency domain decomposition.}
Removing frequency domain decomposition consistently degrades
performance.
For example, the MSE values for Hunan temperature and u-wind increase
from $5.143$ to $5.393$ and from $2.084$ to $2.140$, respectively.
This result shows that separating forecasts into frequency bands reduces
cross-band interference and supports band-specific residual correction.

\myparagraph{Impact of the error correlation prior graph.}
Removing the error correlation prior graph leaves only the static spatial
graph for message passing.
Consequently, the v-wind MSE increases from $2.450$ to $2.534$ on the
Hunan dataset and from $4.501$ to $4.672$ on the French dataset.
This degradation indicates that geographic proximity alone cannot
adequately represent the model-specific and nonlocal dependencies among
forecast errors.

\myparagraph{Impact of the multi-scale temporal module.}
Removing the multi-scale temporal module increases the Hunan temperature
MSE from $5.143$ to $5.671$ and the French u-wind MSE from $4.602$ to
$5.551$.
Figure~\ref{fig:patch_ablation} further shows that the
$\{6,12,24\}$ configuration achieves the lowest MSE and MAE across all
three French regional variables, confirming the benefit of combining
complementary temporal resolutions.

\subsection{Visualization Study}
We visualize the graph structures to examine how the frequency-specific error priors complement the static spatial graph.
Fig.~\ref{fig:full_edge_analysis} compares their edge-distance distributions, structural correlations, and weighted edge overlaps.

\myparagraph{Nonlocal dependencies in error priors.}
Fig.~\ref{fig:full_edge_analysis}(a) shows the edge-distance weighted ECDF over all station pairs.
The spatial graph concentrates its edge weights at short geographic distances, whereas the error prior graphs exhibit heavier tails.
This difference indicates that foundation model errors contain correlated patterns across geographically distant stations, allowing the error graphs to provide nonlocal connections absent from purely spatial message passing.

\myparagraph{Frequency-dependent graph structures.}
Fig.~\ref{fig:full_edge_analysis}(b) and (c) show that the error prior graphs differ across frequency bands and increasingly diverge from the spatial graph.
In particular, their Pearson correlation with the spatial graph decreases from $0.60$ at the low-frequency band to $0.24$ at the high-frequency band.
These results indicate that a single static graph cannot represent error dependencies across all frequency components and support the dynamic fusion of spatial and frequency-specific error graphs.

\section{Conclusion}
\label{sec:conclusion}
We proposed {\sys} to adapt univariate time series foundation models to multi-station weather forecasting.
{\sys} decomposes foundation model forecasts into frequency bands and dynamically integrates band-specific error correlation priors with static spatial relations.
A lightweight graph residual adapter then performs structured cross-station correction while preserving the temporal priors of the
foundation model.
Experiments on regional and global datasets demonstrate consistent improvements over existing baselines.
While this work focuses on interactions among stations, future work will further incorporate dependencies among meteorological variables for joint multivariate and multi-station forecasting.

\section*{AI Use Statement}
Generative AI was used solely for language polishing and grammar correction.
No AI tools were involved in the research process, including methodology design, data analysis, or implementation.

\subsection*{Reproducibility Statement}
The methodological derivations and architectural details required to
reproduce {\sys} are provided in
Appendix~\ref{app:method_details}.
Dataset preprocessing, baseline configurations, hyperparameter settings,
random seeds, hardware, and evaluation protocols are documented in
Appendix~\ref{app:experimental_details}.
Complete MAE results, ablation results, and forecasting cases are
reported in Appendix~\ref{app:additional_results}.
The Supplementary Material includes the core implementation of the
proposed framework. The complete source code, configurations, and data
processing scripts will be publicly released upon acceptance.

\bibliographystyle{plainnat}
\bibliography{main}

@misc{meteonet2020,
  author       = {Gwennaëlle Larvor and Léa Berthomier and Vincent Chabot and Brice Le Pape and Bruno Pradel and Lior Perez},
  title        = {MeteoNet, an open reference weather dataset by METEO FRANCE},
  year         = {2020},
  howpublished = {\url{https://github.com/meteofrance/meteonet}},
  note         = {Accessed: 2025-01-12}
}

@inproceedings{Cao2020StemGNN,
 author = {Cao, Defu and Wang, Yujing and Duan, Juanyong and Zhang, Ce and Zhu, Xia and Huang, Congrui and Tong, Yunhai and Xu, Bixiong and Bai, Jing and Tong, Jie and Zhang, Qi},
 booktitle = {Proceedings of the Advances in Neural Information Processing Systems},
 pages = {17766--17778},
 title = {Spectral Temporal Graph Neural Network for Multivariate Time-series Forecasting},
 volume = {33},
 publisher = {Curran Associates Inc.},
 address = {Red Hook, NY, USA},
 year = {2020}
}

@inproceedings{wu2021autoformer,
  title={Autoformer: Decomposition transformers with auto-correlation for long-term series forecasting},
  author={Wu, Haixu and Xu, Jiehui and Wang, Jianmin and Long, Mingsheng},
  booktitle = {Proceedings of the Advances in Neural Information Processing Systems},
  volume={34},
  pages={22419--22430},
  publisher = {Curran Associates, Inc.},
  address = {Red Hook, NY, USA},
  year={2021}
}

@inproceedings{zhou2022fedformer,
  title = {{FED}former: Frequency Enhanced Decomposed Transformer for Long-term Series Forecasting},
  author =       {Zhou, Tian and Ma, Ziqing and Wen, Qingsong and Wang, Xue and Sun, Liang and Jin, Rong},
  booktitle =    {Proceedings of the International Conference on Machine Learning},
  pages = 	 {27268--27286},
  publisher =   {PMLR},
  address = {Baltimore, MD, USA},
  year = 	 {2022}
}

@article{wu2023Corrformer,
    title={Interpretable weather forecasting for worldwide stations with a unified deep model},
    author={Wu, Haixu and Zhou, Hang and Long, Mingsheng and Wang, Jianmin},
    journal={Nature Machine Intelligence},
    year={2023},
    volume={5},
    number={6},
    pages={602--611},
}

@inproceedings{gerald2024moirai,
title = "Unified Training of Universal Time Series Forecasting Transformers",
author = "Gerald Woo and Chenghao Liu and Akshat Kumar and Caiming Xiong and Silvio Savarese and Doyen Sahoo",
year = "2024",
volume = "235",
pages = "53140--53164",
booktitle = {Proceedings of the International Conference on Machine Learning},
publisher = {PMLR},
address = {Vienna, Austria},
}

@inproceedings{cai2024msgnet,
  title={Msgnet: Learning multi-scale inter-series correlations for multivariate time series forecasting},
  author={Cai, Wanlin and Liang, Yuxuan and Liu, Xianggen and Feng, Jianshuai and Wu, Yuankai},
  booktitle={Proceedings of the AAAI Conference on Artificial Intelligence},
  pages={11141--11149},
  publisher = {AAAI Press},
  address = {Vancouver, BC, Canada},
  year={2024}
}

@article{ansari2024chronos,
title={Chronos: Learning the Language of Time Series},
author={Abdul Fatir Ansari and Lorenzo Stella and Ali Caner Turkmen and Xiyuan Zhang and Pedro Mercado and Huibin Shen and Oleksandr Shchur and Syama Sundar Rangapuram and Sebastian Pineda Arango and Shubham Kapoor and Jasper Zschiegner and Danielle C. Maddix and Hao Wang and Michael W. Mahoney and Kari Torkkola and Andrew Gordon Wilson and Michael Bohlke-Schneider and Bernie Wang},
journal={Transactions on Machine Learning Research},
pages={},
volume={2024},
year={2024},
publisher={Transactions on Machine Learning Research}
}

@inproceedings{das2024timesFM,
author = {Das, Abhimanyu and Kong, Weihao and Sen, Rajat and Zhou, Yichen},
title = {A decoder-only foundation model for time-series forecasting},
year = {2024},
booktitle = {Proceedings of the International Conference on Machine Learning},
publisher = {PMLR},
address = {Vienna, Austria},
pages={10148--10167}
}

@inproceedings{Yang2024FBM,
 author = {Yang, Runze and Cao, Longbing and Li, Jianxun and Yang, Jie},
 booktitle = {Proceedings of the Advances in Neural Information Processing Systems},
 pages = {8515--8540},
 title = {Rethinking Fourier Transform from A Basis Functions Perspective for Long-term Time Series Forecasting},
 volume = {37},
 publisher = {Curran Associates Inc.},
 address = {Red Hook, NY, USA},
 year = {2024}
}

@inproceedings{liu2024timer,
  title = 	 {Timer: Generative Pre-trained Transformers Are Large Time Series Models},
  author =       {Liu, Yong and Zhang, Haoran and Li, Chenyu and Huang, Xiangdong and Wang, Jianmin and Long, Mingsheng},
  booktitle = 	 {Proceedings of the International Conference on Machine Learning},
  pages = 	 {32369--32399},
  year = 	 {2024},
  volume = 	 {235},
  publisher = {PMLR},
  address = {Vienna, Austria},
}

@inproceedings{liu2025sundial,
title={Sundial: A Family of Highly Capable Time Series Foundation Models},
author={Yong Liu and Guo Qin and Zhiyuan Shi and Zhi Chen and Caiyin Yang and Xiangdong Huang and Jianmin Wang and Mingsheng Long},
booktitle = {Proceedings of the International Conference on Machine Learning},
year={2025},
publisher = {PMLR},
pages = {39295--39317},
address = {Vancouver, BC, Canada},
}

@inproceedings{xu2025CDPNet,
author = {Xu, Chujie and Ma, Yuqing and Deng, Haoyuan and Gao, Yajun and Wang, Yudie and Lv, Kai and Liu, Xianglong},
title = {Continuous diffusive prediction network for multi-station weather prediction},
year = {2025},
booktitle = {Proceedings of the International Joint Conference on Artificial Intelligence},
pages = {6714--6722},
publisher = {ijcai.org},
address = {Montreal, Canada},
}

@inproceedings{liu2025timerxl,
title={Timer-{XL}: Long-Context Transformers for Unified Time Series Forecasting},
author={Yong Liu and Guo Qin and Xiangdong Huang and Jianmin Wang and Mingsheng Long},
booktitle={Proceedings of the International Conference on Learning Representations},
year={2025},
pages = {83982--84006},
publisher = {OpenReview.net},
address = {Singapore},
}

@inproceedings{benechehab2025adapts,
title = {{A}da{PTS}: Adapting Univariate Foundation Models to Probabilistic Multivariate Time Series Forecasting},
author = {Benechehab, Abdelhakim and Feofanov, Vasilii and Paolo, Giuseppe and Thomas, Albert and Filippone, Maurizio and K\'{e}gl, Bal\'{a}zs},
booktitle={Proceedings of the International Conference on Machine Learning},
year={2025},
publisher = {PMLR},
pages = {3731--3748},
address = {Vancouver, BC, Canada},
}

@inproceedings{shi2025timemoe,
    title={Time-MoE: Billion-Scale Time Series Foundation Models with Mixture of Experts},
    author={Xiaoming Shi and Shiyu Wang and Yuqi Nie and Dianqi Li and Zhou Ye and Qingsong Wen and Ming Jin},
    booktitle={Proceedings of the International Conference on Learning Representations},
    year={2025},
    pages = {34635--34667},
    publisher = {OpenReview.net},
    address = {Singapore},
}

@misc{noaa2026,
  author       = {{NOAA National Centers for Environmental Information}},
  title        = {Global Hourly - Integrated Surface Database (ISD)},
  year         = {2026},
  howpublished = {\url{https://www.ncei.noaa.gov/products/land-based-station/integrated-surface-database}},
  note         = {Accessed: 2026-01-14}
}

@inproceedings{liu2026beyond,
  title={Beyond Observations: Reconstruction Error-Guided Irregularly Sampled Time Series Representation Learning},
  author={Liu, Jiexi and Cao, Meng and Chen, Songcan},
  booktitle={Proceedings of the AAAI Conference on Artificial Intelligence},
  publisher = {AAAI Press},
  address = {Singapore},
  pages={23712--23720},
  year={2026}
}

@inproceedings{fu2025STELLA,
 author = {Fu, Yisong and Wang, Fei and Shao, Zezhi and Diao, Boyu and Wu, Lin and An, Zhulin and Yu, Chengqing and Li, Yujie and Xu, Yongjun},
 booktitle = {Proceedings of the Advances in Neural Information Processing Systems},
 pages = {82509--82538},
 publisher = {Curran Associates, Inc.},
 title = {On the Integration of Spatial-Temporal Knowledge: A Lightweight Approach to Atmospheric Time Series Forecasting},
 year = {2025}
}

\clearpage
\appendix

\section*{Appendix Overview}

This appendix provides additional methodological derivations,
implementation details, and experimental results that complement the
main paper.
Appendix~\ref{app:method_details} presents the complete frequency
decomposition procedure, details of the Graph Residual Adapter, the
training objective, and a complexity analysis.
Appendix~\ref{app:experimental_details} describes the datasets,
baselines, implementation settings, and evaluation protocol.
Appendix~\ref{app:additional_results} reports the complete MAE results,
MAE ablation results, and representative forecasting cases across the
evaluated datasets, variables, and forecasting horizons.
The notation follows the definitions introduced in the main paper.

\section{Method Details}
\label{app:method_details}

This section expands the operations summarized in
Sections~\ref{sec:dfm} and~\ref{sec:adapter}. We first describe how the
foundation model forecasts are separated into frequency-specific
components. We then provide the complete processing flow of the Graph
Residual Adapter, followed by the optimization objective and
computational complexity.

\subsection{Frequency Domain Decomposition}
\label{app:frequency_decomposition}

The errors produced by a univariate foundation model may exhibit
different dependencies across temporal frequencies. We therefore
decompose its forecasts into low-, mid-, and high-frequency components
before constructing the error correlation prior graphs.

Let $T=T_{\mathrm{out}}$ and let
\begin{equation}
\tilde{\mathbf{x}}^{\mathrm{FM}}_{t:t+T-1}
=
\left[
\tilde{\mathbf{x}}^{\mathrm{FM}}[0],
\ldots,
\tilde{\mathbf{x}}^{\mathrm{FM}}[T-1]
\right]^{\top}
\in\mathbb{R}^{T\times N}
\end{equation}
denote the foundation model forecast over $N$ stations. Here,
$\tilde{\mathbf{x}}^{\mathrm{FM}}[\tau]\in\mathbb{R}^{N}$ is the
forecast at the $\tau$-th step within the output window.

The discrete Fourier transform (DFT) of the forecast is
\begin{equation}
\label{eq:app_dft}
\mathbf{H}[k]
=
\sum_{\tau=0}^{T-1}
\tilde{\mathbf{x}}^{\mathrm{FM}}[\tau]
\exp\!\left(
-j\frac{2\pi k\tau}{T}
\right),
\qquad
k=0,\ldots,T-1,
\end{equation}
where $j=\sqrt{-1}$ and
$\mathbf{H}[k]\in\mathbb{C}^{N}$ contains the Fourier coefficients of
all stations at frequency index $k$.

Because the forecast is real valued, its Fourier spectrum satisfies
the conjugate symmetry property
\begin{equation}
\mathbf{H}[T-k]
=
\overline{\mathbf{H}[k]},
\end{equation}
where $\overline{\mathbf{H}[k]}$ denotes the complex conjugate of
$\mathbf{H}[k]$. It is therefore sufficient to partition the
nonnegative frequency indices
\begin{equation}
\mathcal{K}_{+}
=
\left\{
0,\ldots,\left\lfloor\frac{T}{2}\right\rfloor
\right\}.
\end{equation}

Let
\begin{equation}
K_{+}
=
\left\lfloor\frac{T}{2}\right\rfloor+1
\end{equation}
denote the number of nonnegative frequency indices. We define the
boundaries
\begin{equation}
b_0=0,\qquad
b_1=\left\lceil\frac{K_{+}}{3}\right\rceil,\qquad
b_2=\left\lceil\frac{2K_{+}}{3}\right\rceil,\qquad
b_3=K_{+}.
\end{equation}
and construct three contiguous frequency sets:
\begin{equation}
\begin{aligned}
\mathcal{K}_{\mathrm{low}}
&=
\{b_0,\ldots,b_1-1\},\\
\mathcal{K}_{\mathrm{mid}}
&=
\{b_1,\ldots,b_2-1\},\\
\mathcal{K}_{\mathrm{high}}
&=
\{b_2,\ldots,b_3-1\}.
\end{aligned}
\end{equation}
This construction assigns every nonnegative frequency coefficient to
exactly one band while allowing the three bands to differ in size by
at most one coefficient.

For each frequency band $f\in\mathcal{F}$, where
\begin{equation}
\mathcal{F}
=
\{\mathrm{low},\mathrm{mid},\mathrm{high}\},
\end{equation}
we define a conjugate symmetric binary mask
\begin{equation}
\label{eq:app_frequency_mask}
M_f[k]
=
\begin{cases}
1,
&
k\in\mathcal{K}_f
\text{ or }
(-k\bmod T)\in\mathcal{K}_f,\\
0,
&
\text{otherwise}.
\end{cases}
\end{equation}
The corresponding masked spectrum is
\begin{equation}
\mathbf{H}^{f}[k]
=
M_f[k]\mathbf{H}[k].
\end{equation}

The component associated with frequency band $f$ is reconstructed
using the inverse discrete Fourier transform (IDFT):
\begin{equation}
\label{eq:app_idft}
\tilde{\mathbf{x}}^{\mathrm{FM},f}[\tau]
=
\frac{1}{T}
\sum_{k=0}^{T-1}
\mathbf{H}^{f}[k]
\exp\!\left(
j\frac{2\pi k\tau}{T}
\right).
\end{equation}
The conjugate symmetric masks ensure that each reconstructed component
is real valued. Since the masks form a disjoint partition of the
complete spectrum,
\begin{equation}
\sum_{f\in\mathcal{F}}M_f[k]=1
\end{equation}
for every frequency index $k$. Consequently, the original foundation
model forecast can be recovered exactly:
\begin{equation}
\tilde{\mathbf{x}}^{\mathrm{FM}}[\tau]
=
\sum_{f\in\mathcal{F}}
\tilde{\mathbf{x}}^{\mathrm{FM},f}[\tau].
\end{equation}

We apply the same decomposition to the corresponding ground-truth
sequence when constructing the frequency-specific forecast errors and
training objective. This produces band-specific targets
$\mathbf{x}^{f}_{t:t+T-1}$ that are aligned with
$\tilde{\mathbf{x}}^{\mathrm{FM},f}_{t:t+T-1}$.

\subsection{Graph Residual Adapter Details}
\label{app:adapter_details}

The Graph Residual Adapter performs frequency-specific residual
correction without updating the parameters of the foundation model.
For each $f\in\mathcal{F}$, the adapter receives the band-limited
forecast
$\tilde{\mathbf{x}}^{\mathrm{FM},f}_{t:t+T_{\mathrm{out}}-1}$
and the corresponding dynamic fusion prior graph
$\mathbf{A}^{\mathrm{fusion}}_f$. Its processing flow consists of
multi-scale patch embedding, graph propagation, temporal refinement,
patch unembedding, and scale fusion.

For clarity, the frequency index $f$ is omitted from intermediate
token representations when no ambiguity arises.

\subsubsection{Multi-Scale Time Patch Embedding}

For each patch size $p\in\mathcal{P}$, we first transpose the
band-limited forecast from
$\mathbb{R}^{T_{\mathrm{out}}\times N}$ to
$\mathbb{R}^{N\times T_{\mathrm{out}}}$, placing the station dimension
first. We then divide its temporal dimension into nonoverlapping
segments of length $p$. Let
\begin{equation}
T_p=\frac{T_{\mathrm{out}}}{p}
\end{equation}
denote the resulting number of patch tokens. We define the patching
operation as
\begin{equation}
\operatorname{Reshape}_p(\mathbf{X})
:=
\operatorname{reshape}_{N\times T_p\times p}
\left(\mathbf{X}^{\top}\right).
\end{equation}
Consequently,
\begin{equation}
\operatorname{Reshape}_p
\left(
\tilde{\mathbf{x}}^{\mathrm{FM},f}_{t:t+T_{\mathrm{out}}-1}
\right)
\in\mathbb{R}^{N\times T_p\times p}.
\end{equation}

Each temporal patch is linearly projected into a $D$-dimensional token:
\begin{equation}
\label{eq:app_patch_embedding}
\mathbf{Z}^{(p)}
=
\operatorname{Reshape}_p
\left(
\tilde{\mathbf{x}}^{\mathrm{FM},f}_{t:t+T_{\mathrm{out}}-1}
\right)
\mathbf{W}^{\mathrm{in}}_p
+
\mathbf{b}^{\mathrm{in}}_p,
\end{equation}
where
$\mathbf{W}^{\mathrm{in}}_p\in\mathbb{R}^{p\times D}$ and
$\mathbf{b}^{\mathrm{in}}_p\in\mathbb{R}^{D}$ are learnable
parameters. The resulting representation satisfies
\begin{equation}
\mathbf{Z}^{(p)}
\in\mathbb{R}^{N\times T_p\times D}.
\end{equation}

We repeat this operation for every $p\in\mathcal{P}$. Different patch
sizes provide complementary temporal resolutions: shorter patches
preserve local fluctuations, whereas longer patches provide broader
temporal context for residual correction.

\subsubsection{Graph Normalization and Propagation}

The dynamic fusion graph combines the static spatial graph and the
frequency-specific error correlation graph:
\begin{equation}
\mathbf{A}^{\mathrm{fusion}}_f
=
\mathbf{A}^{\mathrm{spatial}}
+
\alpha_f\mathbf{A}^{\mathrm{error}}_f.
\end{equation}
Although this graph supplies a frequency-specific station dependency
prior, different stations may have different levels of receptivity to
information from neighboring stations. We therefore refine the graph
using a learnable node gating vector
$\mathbf{g}\in\mathbb{R}^{N}$.

In implementation, the node gate is evaluated only on the nonzero edges
of the sparsified fusion graph; the dense notation is used for clarity.
The symmetric node gate is
\begin{equation}
\label{eq:app_node_gate}
\mathbf{G}
=
\operatorname{sig}
\left(
\mathbf{g}\mathbf{1}^{\top}
+
\mathbf{1}\mathbf{g}^{\top}
\right),
\end{equation}
where $\operatorname{sig}(\cdot)$ denotes the sigmoid function and
$\mathbf{1}\in\mathbb{R}^{N}$ is an all-one vector. The gated
adjacency matrix is
\begin{equation}
\overline{\mathbf{A}}^{\mathrm{fusion}}_f
=
\mathbf{A}^{\mathrm{fusion}}_f
\odot
\mathbf{G}.
\end{equation}

Let
\begin{equation}
\mathbf{D}_f
=
\operatorname{diag}
\left(
\overline{\mathbf{A}}^{\mathrm{fusion}}_f
\mathbf{1}
\right)
\end{equation}
be the corresponding row-degree matrix. Row normalization gives
\begin{equation}
\label{eq:app_graph_normalization}
\widetilde{\mathbf{A}}^{\mathrm{fusion}}_f
=
\mathbf{D}_f^{-1}
\overline{\mathbf{A}}^{\mathrm{fusion}}_f.
\end{equation}
This normalization controls the scale of the aggregated messages and
ensures that each row sums to one.

Before propagation, each patch token is normalized over its $D$
features for each sample, station, and patch. We denote its featurewise
mean and standard deviation by
$\boldsymbol{\mu}^{(p)}$ and $\boldsymbol{\sigma}^{(p)}$, respectively,
and define
$\operatorname{TokenNorm}(\mathbf{Z}^{(p)})=
(\mathbf{Z}^{(p)}-\boldsymbol{\mu}^{(p)})/
\max(\boldsymbol{\sigma}^{(p)},10^{-5})$, where the maximum is applied
elementwise.
The statistics are retained to restore the token scale after temporal
refinement. For every temporal token, graph propagation is performed
along the station dimension:
\begin{equation}
\label{eq:app_graph_propagation}
\widetilde{\mathbf{Z}}^{(p)}_{:,\tau,:}
=
\widetilde{\mathbf{A}}^{\mathrm{fusion}}_f
\operatorname{TokenNorm}(\mathbf{Z}^{(p)})_{:,\tau,:},
\qquad
\tau=1,\ldots,T_p.
\end{equation}
Therefore,
\begin{equation}
\widetilde{\mathbf{Z}}^{(p)}
\in\mathbb{R}^{N\times T_p\times D}.
\end{equation}
The propagation operation allows each station to aggregate residual
information from stations connected by geographic relations,
foundation model error dependencies, or both.

\subsubsection{Time Attention and MLP}

Graph propagation operates along the station dimension but does not
explicitly model interactions among temporal patches. We therefore
apply Time Attention independently to each station after graph
propagation. At each patch scale, we first add a learnable positional
embedding $\mathbf{P}^{(p)}\in\mathbb{R}^{1\times T_p\times D}$ to
$\widetilde{\mathbf{Z}}^{(p)}$; the embeddings are initialized to zero.
The temporal module applies pre-normalized attention and an MLP, each
with a residual connection. Its output is rescaled using the token
statistics retained before graph propagation.

For station $i$ and attention head $h$, the query, key, and value
representations are
\begin{equation}
\begin{aligned}
\mathbf{Y}^{(p)}_i
&=
\operatorname{LayerNorm}\!\left(
\widetilde{\mathbf{Z}}^{(p)}_{i,:,:}+\mathbf{P}^{(p)}
\right),\\
\mathbf{Q}^{(p,h)}_i
&=
\mathbf{Y}^{(p)}_i
\mathbf{W}^{Q}_{h},\\
\mathbf{K}^{(p,h)}_i
&=
\mathbf{Y}^{(p)}_i
\mathbf{W}^{K}_{h},\\
\mathbf{V}^{(p,h)}_i
&=
\mathbf{Y}^{(p)}_i
\mathbf{W}^{V}_{h},
\end{aligned}
\end{equation}
where
$\mathbf{W}^{Q}_{h}$,
$\mathbf{W}^{K}_{h}$, and
$\mathbf{W}^{V}_{h}$
are learnable projection matrices.

The temporal attention output of head $h$ is
\begin{equation}
\label{eq:app_time_attention}
\mathbf{U}^{(p,h)}_i
=
\operatorname{softmax}
\left(
\frac{
\mathbf{Q}^{(p,h)}_i
\left(\mathbf{K}^{(p,h)}_i\right)^{\top}
}{
\sqrt{d_h}
}
\right)
\mathbf{V}^{(p,h)}_i,
\end{equation}
where $d_h$ is the dimension of each attention head. The outputs of
all heads are concatenated and projected:
\begin{equation}
\mathbf{S}^{(p)}_i
=
\widetilde{\mathbf{Z}}^{(p)}_{i,:,:}+\mathbf{P}^{(p)}
+
\operatorname{Concat}
\left(
\mathbf{U}^{(p,1)}_i,
\ldots,
\mathbf{U}^{(p,H)}_i
\right)
\mathbf{W}^{O}.
\end{equation}

A lightweight MLP is subsequently applied to obtain
\begin{equation}
\label{eq:app_temporal_mlp}
\begin{aligned}
\mathbf{O}^{(p)}_i
&=\boldsymbol{\mu}^{(p)}_i+
\max(\boldsymbol{\sigma}^{(p)}_i,10^{-5})\odot\\
&\quad\left[\mathbf{S}^{(p)}_i+
\operatorname{MLP}\!\left(
\operatorname{LayerNorm}(\mathbf{S}^{(p)}_i)
\right)\right].
\end{aligned}
\end{equation}
The refined tokens satisfy
$\mathbf{O}^{(p)}\in\mathbb{R}^{N\times T_p\times D}$.
Time Attention captures dependencies across temporal patches, whereas
the MLP refines their feature representations before residual
reconstruction.

For the 24-hour forecasting setting, the branch with
$p=T_{\mathrm{out}}=24$ contains a single token and therefore does not
perform cross-token attention. Instead, this branch provides a
global-window residual representation refined by the projection layers
and MLP, while the shorter patch scales model interactions among
multiple temporal tokens.

\subsubsection{Patch Unembedding}

The refined tokens are mapped back to temporal segments using a
scale-specific output projection:
\begin{equation}
\mathbf{R}^{(p)}
=
\mathbf{O}^{(p)}
\mathbf{W}^{\mathrm{out}}_p
+
\mathbf{b}^{\mathrm{out}}_p,
\end{equation}
where
$\mathbf{W}^{\mathrm{out}}_p\in\mathbb{R}^{D\times p}$ and
$\mathbf{b}^{\mathrm{out}}_p\in\mathbb{R}^{p}$. This gives
\begin{equation}
\mathbf{R}^{(p)}
\in\mathbb{R}^{N\times T_p\times p}.
\end{equation}

The projected segments are rearranged into the original temporal
resolution:
\begin{equation}
\label{eq:app_patch_unembedding}
\Delta\tilde{\mathbf{x}}^{(p,f)}_{t:t+T_{\mathrm{out}}-1}
=
\operatorname{Reshape}^{-1}_p
\left(
\mathbf{R}^{(p)}
\right),
\end{equation}
where
\begin{equation}
\Delta\tilde{\mathbf{x}}^{(p,f)}_{t:t+T_{\mathrm{out}}-1}
\in\mathbb{R}^{T_{\mathrm{out}}\times N}.
\end{equation}

\subsubsection{Dynamic Multi-Scale Fusion}

The usefulness of a temporal scale may vary across samples and
frequency bands. We therefore infer scale weights from the refined
token representations rather than using fixed coefficients.

For each $p\in\mathcal{P}$, global average pooling over the station and
token dimensions gives
\begin{equation}
\label{eq:app_scale_context}
\mathbf{c}^{(p)}
=
\frac{1}{NT_p}
\sum_{i=1}^{N}
\sum_{\tau=1}^{T_p}
\mathbf{O}^{(p)}_{i,\tau,:}
\in\mathbb{R}^{D}.
\end{equation}
A lightweight scoring network produces a scalar score:
\begin{equation}
s^{(p)}
=
\operatorname{MLP}_{\mathrm{score}}
\left(
\mathbf{c}^{(p)}
\right).
\end{equation}
The scores are normalized across patch sizes:
\begin{equation}
\label{eq:app_scale_weight}
\theta^{(p)}
=
\frac{
\exp\left(s^{(p)}\right)
}{
\sum_{q\in\mathcal{P}}
\exp\left(s^{(q)}\right)
}.
\end{equation}
The weights satisfy
\begin{equation}
\theta^{(p)}\geq0,
\qquad
\sum_{p\in\mathcal{P}}\theta^{(p)}=1.
\end{equation}

The residual forecast for frequency band $f$ is
\begin{equation}
\label{eq:app_scale_fusion}
\Delta\tilde{\mathbf{x}}^{f}_{t:t+T_{\mathrm{out}}-1}
=
\sum_{p\in\mathcal{P}}
\theta^{(p)}
\Delta\tilde{\mathbf{x}}^{(p,f)}_{t:t+T_{\mathrm{out}}-1}.
\end{equation}

Finally, a learnable band factor $\beta_f$ controls the correction
magnitude of each frequency component:
\begin{equation}
\label{eq:app_final_forecast}
\hat{\mathbf{x}}_{t:t+T_{\mathrm{out}}-1}
=
\sum_{f\in\mathcal{F}}
\left(
\tilde{\mathbf{x}}^{\mathrm{FM},f}_{t:t+T_{\mathrm{out}}-1}
+
\beta_f
\Delta\tilde{\mathbf{x}}^{f}_{t:t+T_{\mathrm{out}}-1}
\right).
\end{equation}
This residual formulation preserves the temporal forecasts supplied by
the frozen foundation model while introducing structured corrections
based on station dependencies.

\subsection{Training Objective}
\label{app:training_objective}

All foundation model parameters remain frozen during training. The
trainable parameters include the Graph Residual Adapters, node gates
$\mathbf{g}$, graph fusion factors $\alpha_f$, band factors $\beta_f$,
patch projections, Time Attention blocks, MLPs, and scale scorers.

A standard time-domain objective alone may be dominated by
low-frequency components because they typically contain substantially
more signal energy than high-frequency fluctuations. We therefore
combine the MSE of the final forecast with an energy-normalized loss for
each frequency band.

For frequency band $f$, the corrected prediction is
\begin{equation}
\hat{\mathbf{x}}^{f}_{t:t+T_{\mathrm{out}}-1}
=
\tilde{\mathbf{x}}^{\mathrm{FM},f}_{t:t+T_{\mathrm{out}}-1}
+
\beta_f
\Delta\tilde{\mathbf{x}}^{f}_{t:t+T_{\mathrm{out}}-1}.
\end{equation}
Let
$\mathbf{x}^{f}_{t:t+T_{\mathrm{out}}-1}$
denote the corresponding ground-truth frequency component. The overall
training objective is
\begin{equation}
\label{eq:app_training_loss}
\begin{aligned}
\mathcal{L}
&=
\lambda_{\mathrm{mse}}
\operatorname{MSE}
\left(
\hat{\mathbf{x}}_{t:t+T_{\mathrm{out}}-1},
\mathbf{x}_{t:t+T_{\mathrm{out}}-1}
\right)\\
&\quad+
\sum_{f\in\mathcal{F}}
\lambda_f
\sqrt{
\frac{
\left\|
\hat{\mathbf{x}}^{f}_{t:t+T_{\mathrm{out}}-1}
-
\mathbf{x}^{f}_{t:t+T_{\mathrm{out}}-1}
\right\|_{\mathrm{F}}^{2}
}{
\left\|
\mathbf{x}^{f}_{t:t+T_{\mathrm{out}}-1}
\right\|_{\mathrm{F}}^{2}
+
\varepsilon
}
},
\end{aligned}
\end{equation}
where $\|\cdot\|_{\mathrm{F}}$ is the Frobenius norm and
$\varepsilon>0$ is a small constant for numerical stability.

The denominator normalizes each band-specific error by the energy of
its target component. This prevents high-energy low-frequency
components from overwhelming the optimization and maintains meaningful
gradients for the mid- and high-frequency adapters.

\subsection{Complexity Analysis}
\label{app:complexity_analysis}

We analyze the computational and memory complexity of {\sys} by
separating offline graph construction from online adapter computation.
Let $N$ be the number of stations, $T_{\mathrm{out}}$ the forecasting
horizon, $D$ the token dimension, $K$ the number of retained neighbors,
and $T_p=T_{\mathrm{out}}/p$ the number of tokens associated with patch
size $p$.

\subsubsection{Comparison with Joint Spatiotemporal Attention}

A direct approach can flatten the station and time dimensions into
$NT_{\mathrm{out}}$ tokens and apply joint self-attention. Constructing
the corresponding attention matrix requires
\begin{equation}
\mathcal{O}
\left(
N^2T_{\mathrm{out}}^2D
\right)
\end{equation}
time and
\begin{equation}
\mathcal{O}
\left(
N^2T_{\mathrm{out}}^2
\right)
\end{equation}
memory.

A factorized architecture that applies dense spatial attention and
temporal attention separately reduces the time complexity to
\begin{equation}
\mathcal{O}
\left(
N^2T_{\mathrm{out}}D
+
NT_{\mathrm{out}}^2D
\right).
\end{equation}
However, the spatial component remains quadratic in $N$, which becomes
expensive for datasets containing thousands of stations.

\subsubsection{Offline Prior Graph Construction}

The spatial graph depends only on station coordinates and is computed
once before training. Computing all pairwise station distances requires
$\mathcal{O}(N^2)$ operations. The dense spatial similarities are then
reduced to a sparse $K$-nearest-neighbor graph.

For each frequency band, the error correlation graph is constructed
from the training errors. Let $M$ denote the number of error instances
after flattening the training sample and forecast lead-time dimensions.
Computing the full Pearson correlation matrix requires
\begin{equation}
\mathcal{O}(MN^2)
\end{equation}
operations. This computation is performed once for each foundation
model and forecasting setting and does not contribute to the cost of
online forecasting.

After sparsification, each graph contains at most $\mathcal{O}(KN)$
edges. The sparse graph storage cost over all frequency bands is
\begin{equation}
\mathcal{O}
\left(
|\mathcal{F}|KN
\right).
\end{equation}

\subsubsection{Frequency Decomposition}

For each station, computing the DFT over a forecast window of length
$T_{\mathrm{out}}$ requires
$\mathcal{O}(T_{\mathrm{out}}\log T_{\mathrm{out}})$ operations.
The DFT over all stations therefore costs
$\mathcal{O}(NT_{\mathrm{out}}\log T_{\mathrm{out}})$.
Reconstructing the $|\mathcal{F}|$ frequency components using the IDFT
requires
$\mathcal{O}(|\mathcal{F}|NT_{\mathrm{out}}
\log T_{\mathrm{out}})$ operations.
Thus, the total cost of frequency decomposition is
\begin{equation}
\mathcal{O}\!\left(
(1+|\mathcal{F}|)NT_{\mathrm{out}}
\log T_{\mathrm{out}}
\right).
\end{equation}
Storing all band-limited components requires
$\mathcal{O}(|\mathcal{F}|NT_{\mathrm{out}})$ memory.

\subsubsection{Multi-Scale Patch Embedding}

For patch size $p$, patch embedding transforms $NT_p$ temporal patches,
each containing $p$ values, into $D$-dimensional tokens. Its complexity
is
\begin{equation}
\mathcal{O}
\left(
NT_p pD
\right)
=
\mathcal{O}
\left(
NT_{\mathrm{out}}D
\right).
\end{equation}
Patch unembedding has the same asymptotic cost.

\subsubsection{Graph Normalization and Propagation}

The node gate and row normalization operate only on the retained sparse
edges and require $\mathcal{O}(KN)$ operations. Propagating
$T_p$ tokens with dimension $D$ over the sparse graph requires
\begin{equation}
\mathcal{O}
\left(
KNT_pD
\right)
\end{equation}
operations.

This avoids dense spatial message passing, whose complexity would be
$\mathcal{O}(N^2T_pD)$.

\subsubsection{Temporal Refinement}

Time Attention is applied independently to the $T_p$ tokens of each
station. Constructing the temporal attention maps requires
\begin{equation}
\mathcal{O}
\left(
NT_p^2D
\right)
\end{equation}
operations. The query, key, value, output projections, and MLP require
\begin{equation}
\mathcal{O}
\left(
NT_pD^2
\right)
\end{equation}
operations.

Compared with applying temporal attention directly to the
$T_{\mathrm{out}}$ original time steps, patching reduces the quadratic
attention term by a factor of $p^2$ because
$T_p=T_{\mathrm{out}}/p$.

\subsubsection{Scale Scoring and Fusion}

Global pooling requires $\mathcal{O}(NT_pD)$ operations for each patch
scale. The scale scorer operates on a $D$-dimensional context vector,
and its cost is independent of $N$. Reconstructing and combining the
scale-specific residual forecasts requires
$\mathcal{O}(NT_{\mathrm{out}})$ operations.

\subsubsection{Overall Online Complexity}

Combining the above components, the online complexity of the adapter is
\begin{equation}
\label{eq:app_total_complexity}
\mathcal{O}\!\left(
(1+|\mathcal{F}|)NT_{\mathrm{out}}\log T_{\mathrm{out}}
+
|\mathcal{F}|
\sum_{p\in\mathcal{P}}
NT_p
\left(
KD+T_pD+D^2+pD
\right)
\right).
\end{equation}

The corresponding activation memory is dominated by the patch tokens
and temporal attention maps:
\begin{equation}
\mathcal{O}
\left(
|\mathcal{F}|
\sum_{p\in\mathcal{P}}
N
\left(
T_pD+T_p^2
\right)
\right).
\end{equation}
The sparse prior graphs additionally require
$\mathcal{O}(|\mathcal{F}|KN)$ memory.
The band-limited forecasts additionally require
$\mathcal{O}(|\mathcal{F}|NT_{\mathrm{out}})$ memory.

Since $|\mathcal{F}|$, $|\mathcal{P}|$, $K$, $D$, and $T_p$ are fixed
with respect to the number of stations, both the online computation and
activation memory grow linearly with $N$. The proposed adapter
therefore avoids the quadratic station dependence of joint or dense
spatial attention and remains applicable to large multi-station
forecasting datasets.

\section{Experimental Details}
\label{app:experimental_details}

This section provides the information required to reproduce the
experiments. We first describe the three datasets and their
chronological splits, then summarize the baseline configurations,
training settings, random-seed protocol, and evaluation metrics.

\subsection{Dataset Details}
\label{app:dataset_details}

We conduct experiments on three station-based weather datasets with
different geographical scales and station distributions.

\subsubsection{Hunan Regional Dataset}

The Hunan regional dataset is provided by the Hunan Province
Meteorological Bureau. It contains hourly observations from $3{,}570$
weather stations distributed across an approximately
$576\,\mathrm{km}\times576\,\mathrm{km}$ region in Hunan Province,
China.

The dataset covers the period from April 1 to September 30, 2023. Each
station provides temperature at 2 meters and the u- and v-components of
wind at 10 meters. The large number of densely distributed stations
makes this dataset suitable for evaluating scalability and
fine-grained regional station dependencies.

\subsubsection{French Regional Dataset}

The French regional dataset is obtained from
MeteoNet \citep{meteonet2020}, which is released by Meteo France. It
contains hourly observations from $234$ weather stations distributed
within an approximately
$252\,\mathrm{km}\times252\,\mathrm{km}$ region in northwestern France.

The dataset spans from January 1 to December 31, 2016. The evaluated
variables are temperature at 2 meters and the u- and v-components of
wind at 10 meters. Compared with the Hunan dataset, the French dataset
contains fewer and more sparsely distributed stations and therefore
provides a complementary regional forecasting setting.

\subsubsection{Global Dataset}

The global dataset is obtained from the National Centers for
Environmental Information \citep{noaa2026}. It contains hourly
observations from $3{,}850$ stations distributed worldwide between
January 1, 2019, and December 31, 2020.

The evaluated variables are hourly temperature and averaged wind
speed. The global distribution of stations introduces substantially
larger spatial variation than the two regional datasets. 
Because the Global variables were scaled by a factor of 10 during
preprocessing, we divide the resulting MSE by 100 and MAE by 10 before
reporting them.

\subsubsection{Dataset Splits}

For every dataset and variable, observations are arranged
chronologically and divided into training, validation, and test sets
using a $6{:}1{:}3$ ratio. Random shuffling across time is not used.

The training split is used to optimize trainable models and construct
the frequency-specific error correlation prior graphs. The validation
split is used for model selection, and all reported results are
computed on the held-out test split. Neither the validation nor test
targets are used when constructing the error correlation prior graphs.

\subsection{Baseline Details}
\label{app:baseline_details}

We compare {\sys} with three groups of forecasting methods. The grouping
follows the organization used in the Related Work and experimental
sections.

\subsubsection{Time Series Foundation Models}

Timer \citep{liu2024timer} is a generative pretrained time series model
that formulates forecasting through next-token prediction. Sundial
 \citep{liu2025sundial} performs probabilistic forecasting over
continuous-valued observations using flow matching.

We evaluate their officially released pretrained models without
updating their parameters on the target weather datasets. These
experiments measure the zero-shot performance of univariate foundation
models when each station is forecast independently.

\subsubsection{Foundation Models for Multivariate Forecasting}

Timer-XL \citep{liu2025timerxl} extends Timer to multivariate time
series and models dependencies among variables through TimeAttention.
Moirai \citep{gerald2024moirai} is a universal forecasting model
designed to support heterogeneous frequencies and different numbers of
variables.

For Timer-XL and Moirai, we report both zero-shot and fine-tuned
results. Their fine-tuned variants are denoted as Timer-XL-ft and
Moirai-ft, respectively.

AdaPTS \citep{benechehab2025adapts} adapts pretrained univariate
foundation models to multivariate inputs through lightweight
projection adapters. We combine AdaPTS with Timer and Sundial and
denote the resulting variants as AdaPTS (Timer) and AdaPTS (Sundial).
The adapters are trained separately on each target dataset.

\subsubsection{Multi-Station Forecasting Models}

CDPNet \citep{xu2025CDPNet} models continuous spatiotemporal evolution
from discrete station observations. MSGNet \citep{cai2024msgnet}
combines multiscale frequency decomposition with adaptive graph
convolution. Corrformer \citep{wu2023Corrformer} integrates spatial
cross-correlation and temporal autocorrelation for weather
forecasting.

STELLA \citep{fu2025STELLA} incorporates geographic coordinates and
temporal features through spatial-temporal position embedding and uses
a lightweight MLP architecture instead of Transformer layers. These
task-specific models are trained on each target dataset using their
official or recommended configurations.

\subsection{Implementation Details}
\label{app:implementation_details}

Whenever available, all baselines follow their official implementations
and recommended hyperparameter settings. Timer, Sundial, Timer-XL, and
Moirai use their officially released pretrained weights for zero-shot
evaluation. Timer-XL-ft and Moirai-ft are trained on each target
dataset using their official fine-tuning scripts.

For AdaPTS, we evaluate its available adapter configurations and use
the dropout Linear Adapter Ensemble configuration with both Timer and
Sundial. This configuration provides the strongest overall validation
performance across our experimental settings.

We instantiate {\sys} using frozen pretrained Timer and Sundial
backbones. These two variants are denoted as {\sys} (Timer) and
{\sys} (Sundial), respectively. During training, the foundation model
parameters remain frozen, and only the adapter, patch projections,
graph gates, fusion factors, temporal modules, and scale scorers are
updated.

The input window is fixed at $96$ hours to accommodate the input length
required by Timer. The same input window is used for every method to
ensure a consistent comparison. We evaluate forecast horizons
\begin{equation}
T_{\mathrm{out}}\in\{24,96\}.
\end{equation}

The number of frequency bands is fixed to three:
\begin{equation}
\mathcal{F}
=
\{\mathrm{low},\mathrm{mid},\mathrm{high}\}.
\end{equation}
The temporal patch sizes are
\begin{equation}
\mathcal{P}
=
\{6,12,24\}.
\end{equation}
The token dimension is $D=128$. Time Attention uses four heads of
dimension $32$ and an attention dropout rate of $0.1$. The following
MLP has a hidden dimension of $256$ and a dropout rate of $0.1$.
The scale-weight scorer has a hidden dimension of $32$.

Both the spatial correlation graph and each frequency-specific error
correlation prior graph use the same dataset-specific neighborhood size:
$k=32$ for Hunan, $k=16$ for MeteoNet (French regional), and $k=36$
for Global. Self-loops are included in both graph types. 
The training objective uses
$\lambda_{\mathrm{mse}}=1.0$ and
$\lambda_f=0.5$ for every $f\in\mathcal{F}$.

We optimize {\sys} using Adam with an initial learning rate of \(1\times10^{-4}\), which is annealed to \(1\times10^{-6}\) using a cosine schedule, for up to 20 epochs. 
All experiments are
conducted on a single NVIDIA RTX 4090 GPU with 24 GB of memory.

Stochastic experiments are repeated using random seeds $2021$, $2022$,
and $2023$. For each metric, we report the arithmetic mean and sample
standard deviation across the three runs. 

The zero-shot evaluations of Timer and Timer-XL use deterministic
inference and therefore exhibit no variation across random seeds.
Although Sundial and Moirai are not strictly deterministic, their
observed standard deviations under our evaluation protocol are close
to zero and negligible at the reported precision. We therefore report
these four foundation models without standard deviations. For all
trainable models, results are reported as the mean and sample standard
deviation over three random seeds.

\subsection{Evaluation Metrics}
\label{app:evaluation_metrics}

We evaluate forecasting accuracy using mean squared error (MSE) and
mean absolute error (MAE). Lower values indicate better performance for
both metrics.

Let $\Omega$ denote the set of evaluated sample, forecast-step, and
station indices. MSE is defined as
\begin{equation}
\operatorname{MSE}
=
\frac{1}{|\Omega|}
\sum_{(s,\tau,i)\in\Omega}
\left(
\hat{x}_{s,\tau,i}
-
x_{s,\tau,i}
\right)^2,
\end{equation}
and MAE is defined as
\begin{equation}
\operatorname{MAE}
=
\frac{1}{|\Omega|}
\sum_{(s,\tau,i)\in\Omega}
\left|
\hat{x}_{s,\tau,i}
-
x_{s,\tau,i}
\right|.
\end{equation}

Let $m_r$ denote the metric obtained using the $r$-th random seed. For
$R=3$ runs, the reported mean is
\begin{equation}
\bar{m}
=
\frac{1}{R}
\sum_{r=1}^{R}m_r,
\end{equation}
and the sample standard deviation is
\begin{equation}
s_m
=
\sqrt{
\frac{1}{R-1}
\sum_{r=1}^{R}
\left(
m_r-\bar{m}
\right)^2
}.
\end{equation}
The standard deviation therefore uses Bessel's correction, corresponding
to \texttt{ddof=1} in numerical implementations.

\subsubsection{Weighted Edge Overlap}
\label{app:weighted_edge_overlap}
For Fig.~\ref{fig:full_edge_analysis}(c), we compare the spatial
and error prior graphs after top-$k$ sparsification and
symmetrization. We exclude self-loops and assign zero weight to
absent edges.
For two graphs $A$ and $B$, let
$\mathcal{E}_{AB}=\{(i,j):i<j,\ A_{ij}\text{ and }B_{ij}\text{ are finite}\}$
and $w^A_{ij}=\max(A_{ij},0)$, $w^B_{ij}=\max(B_{ij},0)$.
Their weighted edge overlap (weighted Jaccard index) is
\begin{equation}
\label{eq:weighted_edge_overlap}
J(A,B)=
\frac{\sum_{(i,j)\in\mathcal{E}_{AB}}\min(w^A_{ij},w^B_{ij})}
{\sum_{(i,j)\in\mathcal{E}_{AB}}\max(w^A_{ij},w^B_{ij})}.
\end{equation}
The overlap is undefined if the denominator is zero.

\section{Additional Experimental Results}
\label{app:additional_results}
This section supplements the main experiments with complete MAE results
and qualitative forecasting cases. The MAE results cover all datasets,
variables, and forecasting horizons under the same experimental settings
as the MSE results reported in the main paper. The forecasting cases
further illustrate the behavior of different methods on representative
weather sequences.

\subsection{Complete MAE Results}
\label{app:mae_results}
Table~\ref{tab:all_mae} reports the complete MAE results on the Hunan
regional, French regional, and global datasets for forecasting horizons
of 24 and 96 hours. 
For trainable models, we report the mean and sample standard deviation
over three runs. Timer and Timer-XL use deterministic inference.
Sundial and Moirai are not strictly deterministic, but their observed
variations are negligible relative to their mean values; their standard
deviations are therefore omitted. Best and second-best results are
marked in bold and underlined, respectively, based on the unrounded means.

Table~\ref{tab:ablation_mae} provides the complete MAE results
corresponding to the MSE ablation study in the main paper. The complete
model achieves the lowest MAE across all variables on both regional
datasets, confirming that the improvements observed in MSE are also
consistent under the absolute-error metric.
Removing either the frequency domain decomposition or the error
correlation prior graph consistently degrades the forecasting accuracy.
The variant without the multi-scale temporal module exhibits the largest
overall degradation, particularly for temperature and wind forecasting
on the French regional dataset. These results further demonstrate that
frequency-specific correction, error-aware spatial propagation, and
multi-scale temporal modeling contribute complementary improvements.

\begin{table}[t]
\centering
\caption{Complete MAE results on the Hunan regional, French regional,
and global datasets with a 96-hour input window and forecast horizons
of 24 and 96 hours. Results are reported as the mean and sample standard
deviation over three random seeds where applicable. Best and second-best
results are shown in \best{bold} and \second{underlined}, respectively.
}
\label{tab:all_mae}

\begingroup
\scriptsize
\renewcommand{\arraystretch}{1.45}
\setlength{\tabcolsep}{0.8pt}

\resizebox{\linewidth}{!}{%
\begin{tabular}{@{}c|c|c|cc!{\vrule width 1.2pt}cccccccccccc@{}}
\hline
\multicolumn{3}{c|}{Models}
& \makecell{\sys\\(Timer)}
& \makecell{\sys\\(Sundial)}
& Timer
& Sundial
& \makecell{AdaPTS\\(Timer)}
& \makecell{AdaPTS\\(Sundial)}
& Moirai$_{\mathrm{ft}}$
& Moirai
& Timer-XL$_{\mathrm{ft}}$
& Timer-XL
& STELLA
& CDPNet
& MSGNet
& Corrformer \\
\hline

\multirow{6}{*}{\rotatebox{90}{Hunan}}
& \multirow{2}{*}{\rotatebox{90}{temp}}
& 24
& \best{1.552}\std{0.031}
& \second{1.586}\std{0.005}
& 1.627
& 1.679
& 1.661\std{0.055}
& 1.735\std{0.056}
& 1.715\std{0.103}
& 1.590
& 1.697\std{0.015}
& 1.662
& 2.236\std{0.131}
& 2.023\std{0.111}
& 1.760\std{0.110}
& 1.841\std{0.100} \\
&
& 96
& \second{1.952}\std{0.016}
& \best{1.947}\std{0.005}
& 1.987
& 2.119
& 2.094\std{0.078}
& 2.861\std{0.127}
& 1.981\std{0.019}
& 1.991
& 2.269\std{0.041}
& 1.966
& 3.100\std{0.181}
& 4.027\std{0.564}
& 2.680\std{0.095}
& 2.256\std{0.068} \\
\cline{2-17}

& \multirow{2}{*}{\rotatebox{90}{v-wind}}
& 24
& \best{0.737}\std{0.003}
& 0.761\std{0.002}
& \second{0.745}
& 0.817
& 0.751\std{0.032}
& 0.763\std{0.020}
& 0.766\std{0.012}
& 0.771
& 0.794\std{0.002}
& 0.766
& 0.759\std{0.011}
& 0.797\std{0.013}
& 0.775\std{0.054}
& 0.882\std{0.064} \\
&
& 96
& \best{0.842}\std{0.004}
& 0.931\std{0.028}
& 0.850
& 0.899
& 0.875\std{0.040}
& 0.882\std{0.044}
& 0.891\std{0.039}
& 0.898
& 0.847\std{0.040}
& \second{0.846}
& 0.888\std{0.045}
& 1.114\std{0.335}
& 1.001\std{0.039}
& 0.949\std{0.073} \\
\cline{2-17}

& \multirow{2}{*}{\rotatebox{90}{u-wind}}
& 24
& 0.689\std{0.002}
& \best{0.661}\std{0.018}
& 0.676
& 0.751
& 0.698\std{0.023}
& 0.712\std{0.016}
& 0.711\std{0.003}
& 0.710
& 0.698\std{0.003}
& 0.686
& \second{0.674}\std{0.010}
& 0.711\std{0.017}
& 0.719\std{0.038}
& 0.676\std{0.045} \\
&
& 96
& \second{0.696}\std{0.004}
& \best{0.693}\std{0.018}
& 0.709
& 0.795
& 0.717\std{0.022}
& 0.740\std{0.011}
& 0.767\std{0.005}
& 0.778
& 0.699\std{0.003}
& 0.720
& 0.707\std{0.006}
& 0.769\std{0.038}
& 0.792\std{0.042}
& 0.724\std{0.050} \\
\hline

\multirow{6}{*}{\rotatebox{90}{French}}
& \multirow{2}{*}{\rotatebox{90}{temp}}
& 24
& 1.943\std{0.011}
& \second{1.927}\std{0.011}
& 2.071
& 2.107
& 2.013\std{0.087}
& 2.073\std{0.042}
& 1.971\std{0.011}
& 2.032
& \best{1.925}\std{0.037}
& 2.009
& 2.230\std{0.031}
& 3.703\std{0.307}
& 2.013\std{0.096}
& 2.452\std{0.078} \\
&
& 96
& \best{2.692}\std{0.026}
& 2.771\std{0.022}
& \second{2.721}
& 2.809
& 2.754\std{0.035}
& 2.770\std{0.052}
& 2.743\std{0.127}
& 2.734
& 2.850\std{0.201}
& 2.768
& 3.380\std{0.161}
& 7.446\std{0.104}
& 2.843\std{0.222}
& 2.824\std{0.073} \\
\cline{2-17}

& \multirow{2}{*}{\rotatebox{90}{v-wind}}
& 24
& \best{1.487}\std{0.006}
& \second{1.498}\std{0.002}
& 1.655
& 1.724
& 1.625\std{0.065}
& 1.582\std{0.075}
& 1.592\std{0.016}
& 1.696
& 1.663\std{0.007}
& 1.663
& 1.654\std{0.014}
& 1.596\std{0.051}
& 1.573\std{0.114}
& 1.731\std{0.134} \\
&
& 96
& \best{1.821}\std{0.005}
& 1.868\std{0.072}
& 2.007
& 2.147
& 1.943\std{0.115}
& \second{1.863}\std{0.044}
& 2.043\std{0.014}
& 2.279
& 2.051\std{0.036}
& 2.009
& 1.981\std{0.057}
& 1.962\std{0.077}
& 2.140\std{0.163}
& 2.071\std{0.096} \\
\cline{2-17}

& \multirow{2}{*}{\rotatebox{90}{u-wind}}
& 24
& \best{1.556}\std{0.007}
& \second{1.560}\std{0.001}
& 1.820
& 1.934
& 1.624\std{0.024}
& 1.700\std{0.024}
& 1.686\std{0.011}
& 1.758
& 1.583\std{0.008}
& 1.854
& 1.776\std{0.018}
& 1.648\std{0.139}
& 1.617\std{0.066}
& 1.809\std{0.109} \\
&
& 96
& \best{2.148}\std{0.006}
& \second{2.149}\std{0.028}
& 2.327
& 2.524
& 2.169\std{0.114}
& 2.154\std{0.026}
& 2.350\std{0.032}
& 2.562
& 2.188\std{0.050}
& 2.336
& 2.242\std{0.038}
& 2.254\std{0.051}
& 2.320\std{0.130}
& 2.303\std{0.114} \\
\hline

\multirow{4}{*}{\rotatebox{90}{Global}}
& \multirow{2}{*}{\rotatebox{90}{temp}}
& 24
& 1.903\std{0.003}
& \second{1.881}\std{0.004}
& 2.143
& 2.186
& 2.050\std{0.086}
& 2.034\std{0.120}
& 2.052\std{0.003}
& 2.109
& 1.949\std{0.036}
& 2.337
& \best{1.861}\std{0.003}
& 2.126\std{0.020}
& 2.251\std{0.116}
& 1.946\std{0.063} \\
&
& 96
& \best{2.625}\std{0.010}
& \second{2.635}\std{0.040}
& 2.886
& 3.003
& 2.788\std{0.088}
& 2.857\std{0.093}
& 2.785\std{0.030}
& 2.998
& 2.757\std{0.015}
& 3.089
& 2.723\std{0.014}
& 2.725\std{0.065}
& 3.024\std{0.031}
& 2.914\std{0.048} \\
\cline{2-17}

& \multirow{2}{*}{\rotatebox{90}{wind}}
& 24
& \second{1.282}\std{0.001}
& \best{1.278}\std{0.001}
& 1.377
& 1.494
& 1.375\std{0.051}
& 1.398\std{0.011}
& 1.371\std{0.004}
& 1.449
& 1.368\std{0.013}
& 1.405
& 1.290\std{0.001}
& 1.408\std{0.011}
& 1.564\std{0.057}
& 1.366\std{0.094} \\
&
& 96
& \second{1.488}\std{0.036}
& 1.500\std{0.024}
& 1.563
& 1.691
& 1.532\std{0.072}
& 1.544\std{0.055}
& 1.560\std{0.007}
& 1.763
& 1.497\std{0.004}
& 1.578
& \best{1.476}\std{0.004}
& 1.532\std{0.027}
& 1.802\std{0.008}
& 1.559\std{0.051} \\
\hline
\end{tabular}%
}

\endgroup
\end{table}

\begin{table}[t]
\centering
\caption{Complete MAE results of the ablation study on the Hunan
regional and French regional datasets. The input window is 96 hours,
and the forecast horizon is 24 hours. Results are reported as the mean
and sample standard deviation over three random seeds. Best results are
shown in \textbf{bold}.}
\label{tab:ablation_mae}

\begingroup
\small
\renewcommand{\arraystretch}{1}
\setlength{\tabcolsep}{3pt}

\begin{tabular}{@{}lcccccc@{}}
\toprule
\multirow{2}{*}{Model variant}
& \multicolumn{3}{c}{Hunan regional}
& \multicolumn{3}{c}{French regional} \\
\cmidrule(lr){2-4}
\cmidrule(lr){5-7}
& temp & v-wind & u-wind
& temp & v-wind & u-wind \\
\midrule

\textbf{\sys{} (Timer)}
& \best{1.552}\std{0.031}
& \best{0.737}\std{0.003}
& \best{0.689}\std{0.002}
& \best{1.943}\std{0.011}
& \best{1.487}\std{0.006}
& \best{1.556}\std{0.007} \\

\midrule

w/o frequency domain decomposition
& 1.682\std{0.041}
& 0.791\std{0.008}
& 0.712\std{0.016}
& 1.985\std{0.020}
& 1.586\std{0.016}
& 1.641\std{0.004} \\

w/o error correlation prior graph
& 1.684\std{0.038}
& 0.793\std{0.003}
& 0.713\std{0.004}
& 1.977\std{0.022}
& 1.619\std{0.004}
& 1.636\std{0.018} \\

w/o multi-scale temporal module
& 1.753\std{0.062}
& 0.797\std{0.008}
& 0.714\std{0.010}
& 2.002\std{0.014}
& 1.652\std{0.010}
& 1.737\std{0.015} \\

\bottomrule
\end{tabular}

\endgroup
\end{table}

\subsection{Forecasting Case Studies}
\label{app:forecasting_cases}

To complement the aggregate evaluation metrics, we present representative
forecasting cases selected from the test sets. The cases cover different
datasets, weather variables, and forecasting horizons, including both
smooth temporal trends and rapid local fluctuations. For each case, we
compare the ground-truth observations with the predictions produced by
the frozen foundation model and {\sys}.

These comparisons provide a direct view of how the proposed adapter
modifies the original foundation model forecasts. In particular, they
allow us to examine whether {\sys} can correct systematic forecast
deviations, preserve the overall temporal trend, and capture local
variations across correlated stations. We also include long-horizon
cases to investigate the accumulation of forecasting errors as the
prediction horizon increases.

\begin{figure}[t]
    \centering
    \includegraphics[width=\textwidth]{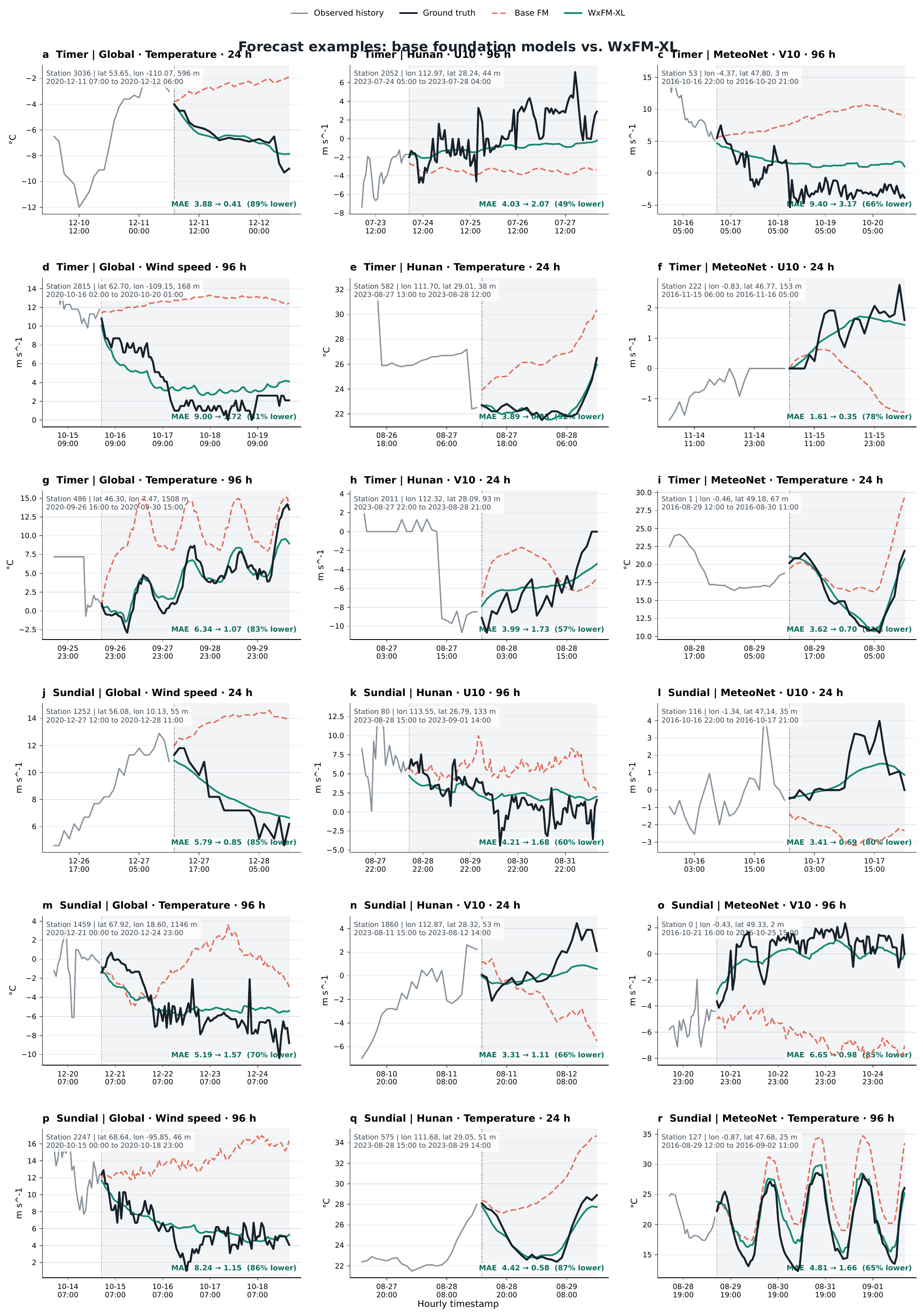}
    \caption{Representative forecasting cases on the evaluated
    datasets. The plots compare the ground truth with predictions from the frozen foundation model and {\sys}.}
    \label{fig:forecasting_cases}
\end{figure}

\end{document}